\makeatletter
\def\input@path{{./}{arxiv/}}
\makeatother
\documentclass{article} 
\usepackage{iclr2026_conference,times}

\usepackage{amsmath,amsfonts,bm}

\def\eqref#1{equation~\ref{#1}}

\def\1{\bm{1}}

\DeclareMathAlphabet{\mathsfit}{\encodingdefault}{\sfdefault}{m}{sl}
\SetMathAlphabet{\mathsfit}{bold}{\encodingdefault}{\sfdefault}{bx}{n}

\usepackage[utf8]{inputenc} 
\usepackage[T1]{fontenc}    
\usepackage{pifont}
\usepackage{colortbl}
\usepackage{tabularx}
\usepackage{xcolor}
\usepackage[most]{tcolorbox}
\usepackage{hyperref}       
\usepackage{url}            
\usepackage{booktabs}       
\usepackage{amsfonts}       
\usepackage{amsmath}        
\usepackage{nicefrac}       
\usepackage{microtype}      
\usepackage{graphicx}       
\usepackage{array}         
\usepackage{longtable}     
\usepackage{algorithm}
\usepackage{algpseudocode}

\title{Chain Of Interaction Benchmark (COIN): When Reasoning meets Embodied Interaction}

\author{
  Xianhao Wang$^{1,2}$ \And
  Xiaojian Ma$^2$ \And
  Haozhe Hu$^3$ \And
  Rongpeng Su$^{1,2}$ \And
  Yutian Cheng$^{4,2}$ \And
  Zhou Ziheng$^{5,2}$ \And
  Hangxin Liu$^2$ \And
  Lei Liu$^1$ \And
  Bin Li$^1$ \And
  Qing Li$^2$ \\
  $^1$University of Science and Technology of China \\
  $^2$Beijing Institute for General Artificial Intelligence (BIGAI) \\
  $^3$Xidian University\\
  $^4$Shanghai Jiao Tong University\\
  $^5$University of California, Los Angeles
}

\makeatletter
\DeclareRobustCommand\onedot{\futurelet\@let@token\@onedot}
\def\@onedot{\ifx\@let@token.\else.\null\fi\xspace}

\makeatother

\newcommand{\cmark}{\ding{51}} 
\newcommand{\xmark}{\ding{55}} 
\iclrfinalcopy 
\begin{document}

\maketitle

\begin{abstract}
Generalist embodied agents must perform interactive, causally-dependent reasoning, continually interacting with the environment, acquiring information, and updating plans to solve long-horizon tasks before they could be adopted in real-life scenarios. For instance, retrieving an apple from a cabinet may require opening multiple doors and drawers before the apple becomes visible and reachable—demanding sequential interaction under partial observability. However, existing benchmarks fail to systematically evaluate this essential capability. We introduce \textbf{COIN}, a benchmark designed to assess interactive reasoning in realistic robotic manipulation through three key contributions. First, we construct \textbf{COIN-50}: 50 interactive tasks in daily scenarios, and create \textbf{COIN-Primitive} required by causally-dependent tasks, and \textbf{COIN-Composition} with mid-term complexity for skill learning and generalization evaluation. Second, we develop a low-cost mobile AR teleoperation system and collect the COIN-Primitive Dataset with 50 demonstrations per primitive task (1,000 in total). Third, we develop systematic evaluation metrics about execution stability and generalization robustness to evaluate \textbf{CodeAsPolicy}, \textbf{VLA}, and language-conditioned \textbf{H-VLA} approaches. Our comprehensive evaluation reveals critical limitations in current methods: models struggle with interactive reasoning tasks due to significant gaps between visual understanding and motor execution. We provide fine-grained analysis of these limitations.
\end{abstract}

\section{Introduction}

Recent advances in large-scale pretraining \cite{nvidia_gr00t_2025, openpi, brohan_rt-1_2023} and the creation of diverse datasets \cite{OXE, droid} and benchmarks \cite{zhang_vlabench_nodate, li2024evaluating, liu_libero_2023} have significantly advanced robotic manipulation capabilities. However, current benchmarks primarily focus on simplified tasks that fail to capture the complexity of real-world manipulation challenges, particularly those requiring interaction and causal reasoning over long time horizons in partially observable environments.

Consider a robot tasked with "open a locked door". This seemingly simple instruction requires a sequence of interdependent actions: locating the keyhole, inserting and turning the key, and then rotating the handle with trials for the right directions. Such tasks demand what we term \textbf{interactive reasoning}—the ability to continually interact with the environment, gather information, update beliefs, and adapt plans accordingly. This requires multiple capabilities: perceiving partial environmental states, reasoning about causal dependencies between actions, maintaining memory of previous interactions, and dynamically adjusting strategies based on feedback. This capability remains beyond the reach of most current Vision-Language-Action (VLA) models and VLM-based planning approaches.

To address this gap, we introduce \textbf{COIN} (Chain Of INteraction) Benchmark, consisting of three complementary components: \textbf{COIN-50}, featuring 50 interactive reasoning tasks grounded in everyday activities (with one demonstration per task); \textbf{COIN-Primitive}, containing 20 fundamental manipulation skills that serve as building blocks (with approximately 50 trajectories per task); and \textbf{COIN-Composition}, bridging \textbf{COIN-Primitive} and \textbf{COIN-50} for evaluating the robustness of VLA learning across visual understanding and instruction variations. Unlike previous benchmarks that primarily test perception or simple manipulation, our tasks are systematically organized according to a taxonomy of reasoning capabilities required in partially observable environments. Based on our analysis, we categorize these capabilities into three principal domains: (1) \textbf{Object-Centric Reasoning}, encompassing physical property inference, spatial reasoning, mechanism understanding, and visual reasoning; (2) \textbf{Robot-Centric Reasoning}, covering control optimization and embodiment awareness (such as collision handling); and (3) \textbf{Compositional Reasoning}, including tool-mediated problem solving, failure-driven adaptation, hierarchical planning, and experience utilization. These capabilities, essential for robots to function effectively in human environments, remain underexplored in existing benchmarks.

To support algorithm development and evaluation, we created a low-cost, phone-based teleoperation system (hardware cost under \$20 according to second-hand websites in China) inspired by \cite{ayyan2024mujocoar}. Using this system, we collected the COIN-Primitive Dataset—over 1,000 expert demonstrations across 20 fundamental manipulation skills recorded from multiple viewpoints. These primitives serve as essential building blocks for VLA model fine-tuning and compositional task solving. \textbf{Our contributions include:}
\vspace{-3mm}
\begin{figure}
\centering
\includegraphics[width=0.95\linewidth]{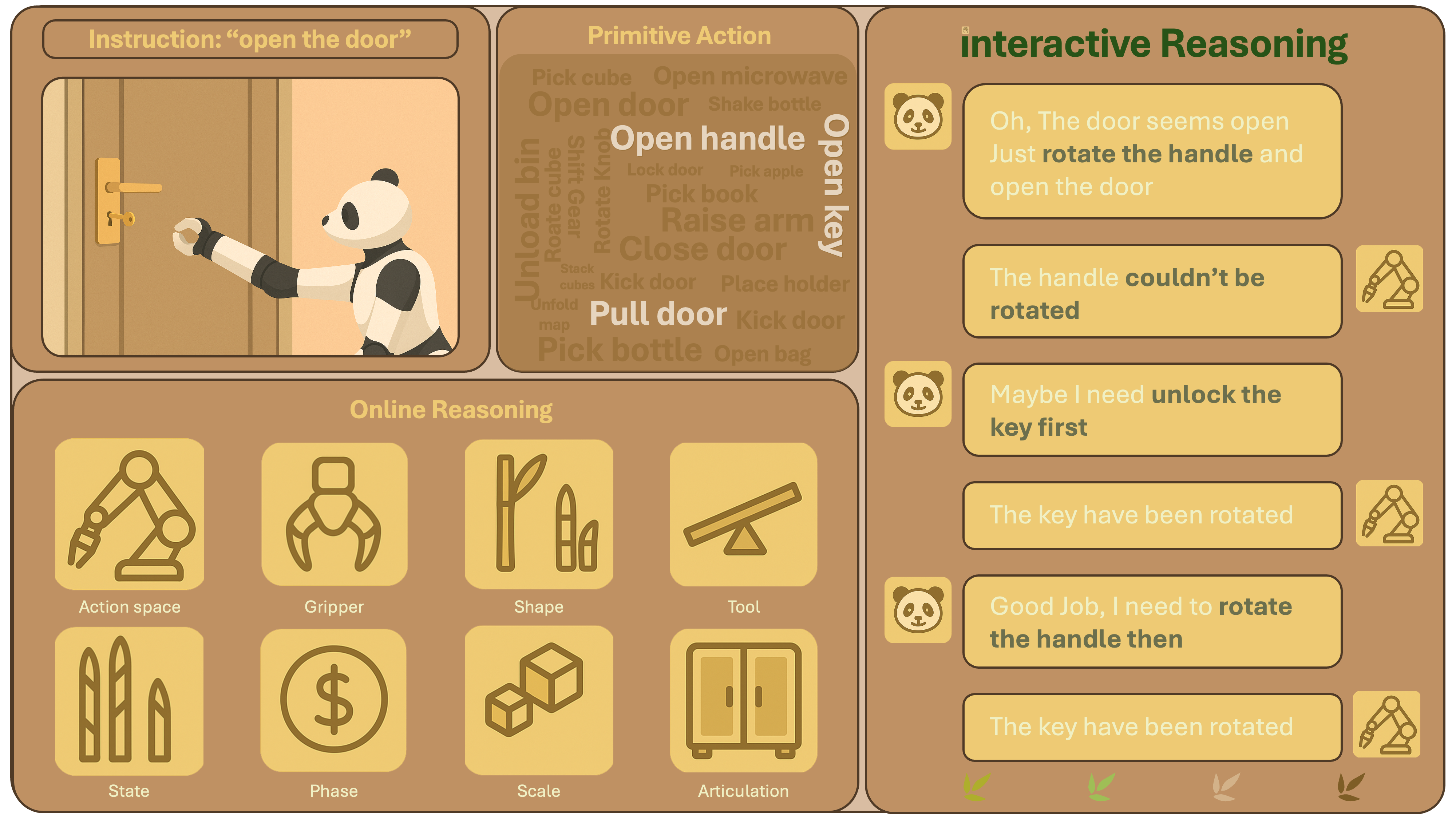}
\caption{\label{fig:teaser}An illustration of \textbf{COIN}. Our benchmark focuses on evaluating the crucial \textbf{interactive reasoning} ability of Vision-Language-Action (VLA) models and VLM-based robotic planning systems, covering both rich reasoning knowledge and diverse primitive actions.}
\end{figure}
\setlength{\leftmargini}{1.2em}
\begin{enumerate}
\item \textbf{COIN Benchmark}: We construct \textbf{COIN-50} with 50 interactive tasks in daily scenarios, \textbf{COIN-Primitive} with 20 fundamental manipulation skills required by causally-dependent tasks, and \textbf{COIN-Composition} with mid-term complexity for skill learning and generalization evaluation, systematically organized according to a principled taxonomy of interactive reasoning capabilities.

\item \textbf{Low-Cost Mobile AR Teleoperation System and Dataset}: We develop a smartphone-based teleoperation system (hardware cost under \$20 according to second-hand websites in China) and collect the COIN-Primitive Dataset with 50 demonstrations per primitive task (1,000 in total), enabling accessible data collection for the robotics community.

\item \textbf{Systematic Evaluation Metrics and Analysis}: We develop comprehensive evaluation metrics about execution stability and generalization robustness to evaluate \textbf{CodeAsPolicy}, \textbf{VLA}, and \textbf{H-VLA} approaches, revealing critical limitations including significant gaps between visual understanding and motor execution, and provide fine-grained analysis for these limitations.
\end{enumerate}
\vspace{-3mm}

\section{Related Works}
\subsection{Vision-Language-Action Models and Approaches}

\textbf{CodeAsPolicy} approaches~\cite{code_as_policy} combine VLMs with predefined skills to orchestrate perception modules \cite{sam, sam2, som} and low-level controllers in a modular, zero-shot framework. Works like~\cite{copa, voxposer, huang_rekep_2024} excel in generalization but struggle with online adaptation due to their ``plan-then-execute'' paradigm, where VLMs disengage after initial planning. While recent work~\cite{duan_manipulate-anything_2024} introduces replanning mechanisms, significant challenges remain in dynamic, partially observable scenarios requiring continuous interactive reasoning.

\textbf{End-to-End VLA models}~\cite{brohan_rt-1_2023, rt2, nvidia_gr00t_2025, li_cogact_2024} directly map visual observations and language to robotic actions via token prediction, learning policies through imitation. Their unified architecture enables emergent reasoning through large-scale pretraining. Despite success in basic manipulation tasks, these models struggle with long-horizon scenarios requiring state maintenance and adaptive planning over extended interactions.

\textbf{Hierarchical VLA (H-VLA) architectures}~\cite{helix, geminiroboticsteam2025geminiroboticsbringingai} bridge planning and execution by decomposing high-level instructions into subtasks coordinated with low-level executors. This approach combines explicit reasoning with learned behaviors, showing promise in complex manipulation tasks and advancing toward more generalist robotic systems.

\subsection{Robot Manipulation Benchmarks}

\begin{table}[t!]
\centering
\footnotesize
\vspace{-3mm}
\begin{tabularx}{\textwidth}{l*{2}{>{\centering\arraybackslash}X}*{2}{>{\centering\arraybackslash}X}*{5}{>{\columncolor{gray!10}\centering\arraybackslash}X}}
\toprule
\textbf{Benchmark} & \textbf{Tasks} & \textbf{Demos} & \textbf{Avg.\newline Steps} & \textbf{Cont.\newline Action} & \textbf{\cellcolor{gray!10}Caus. \newline Dep.} & \textbf{\cellcolor{gray!10}Visual\newline Comp.} & \textbf{\cellcolor{gray!10}Inter.\newline Reas.} & \textbf{\cellcolor{gray!10}Visual\newline Unobs.} & \textbf{\cellcolor{gray!10}Mech.\newline Unobs.} \\
\midrule

\multicolumn{5}{l}{\textbf{Robot Manipulation Benchmarks}} & \multicolumn{5}{c}{} \\

CALVIN~\cite{mees_calvin_2022}          & 34 & 200K & 30 & \cmark & \xmark & \cmark & \xmark & \xmark & \xmark \\
Arnold~\cite{arnold}           & 8 & 40 & 125.8 & \cmark & \xmark & \xmark & \xmark & \xmark & \xmark \\
SimplerEnv~\cite{zhu_robosuite_nodate}       & 10 & N/A & 52.3 & \cmark & \xmark & \xmark & \xmark & \xmark & \xmark \\
Libero~\cite{liu_libero_2023}           & 130 & 50/task & 77.3 & \cmark & \cmark & \cmark & \xmark & \xmark & \xmark \\
VLABench~\cite{zhang_vlabench_nodate}         & 100 & 163 & 157.2 & \cmark & \xmark & \cmark & \xmark & \xmark & \xmark \\
RoboCASA~\cite{zheng_robocas_2024}     & 100 & 100/task & 371.9 & \cmark & \cmark & \cmark & \xmark & \cmark & \cmark \\
EmbodiedBench~\cite{yang_embodiedbench_2025}    & 100 & N/A & N/A & \xmark & \xmark & \cmark & \xmark & \cmark & \xmark \\
RoboVerse~\cite{robotverse}        & 1000 & 9331 & N/A & \cmark & \cmark & \cmark & \xmark & \cmark & \cmark \\
\midrule
\multicolumn{5}{l}{\textbf{Vision-Language Reasoning Benchmarks}} & \multicolumn{5}{c}{} \\
VLMbench~\cite{li_evaluating_2024}         & 100 & N/A & N/A & \xmark & \cmark & \cmark & \xmark & \xmark & \xmark \\
ClevrSkills~\cite{haresh_clevrskills_2024}      & 12 & N/A & N/A & \xmark & \cmark & \cmark & \xmark & \xmark & \xmark \\
ReflectVLM~\cite{feng_reflective_2025}       & 50 & N/A & N/A & \xmark & \cmark & \cmark & \cmark & \xmark & \xmark \\
\midrule
\rowcolor{gray!20}
\textbf{COIN (Ours)} & \textbf{90} & \textbf{1000+} & \textbf{988.9} & \cmark & \cmark & \cmark & \textbf{\cmark} & \cmark & \cmark \\
\bottomrule
\end{tabularx}
\caption{Comprehensive benchmark comparison including quantitative metrics and reasoning capabilities. COIN demonstrates the longest average trajectory length (988.9 steps) and uniquely combines all critical reasoning capabilities, particularly interactive reasoning. Our systematic evaluation framework with 1000+ demonstrations across 90 tasks provides unprecedented depth for analyzing interactive manipulation.}
\label{tab:benchmark_comparison}
\end{table}
\vspace{-3mm}

\textbf{Robot manipulation benchmarks} excel in physical interaction and control capabilities. Works like Arnold~\cite{arnold} and SimplerEnv~\cite{zhu_robosuite_nodate} offer photorealistic simulation but lack reasoning components. Libero~\cite{liu_libero_2023} and RoboCASA~\cite{zheng_robocas_2024} incorporate partial observability, but most lack the combination of dynamic interaction, failure recovery, and interactive reasoning needed for realistic scenarios. Table~\ref{tab:benchmark_comparison} shows that these benchmarks do not cover interactive reasoning well, while our benchmark emphasizes this crucial capability of embodied AI.

\textbf{Vision-language embodied reasoning benchmarks} prioritize reasoning over physical realism. VLMbench~\cite{li_evaluating_2024} and ClevrSkills~\cite{haresh_clevrskills_2024} support causal reasoning in simplified environments, while ReflectVLM~\cite{feng_reflective_2025} offers failure recovery but limited physical interaction. COIN uniquely bridges this gap by combining all eight critical dimensions shown in Table~\ref{tab:benchmark_comparison}, enabling evaluation of true interactive reasoning in realistic, partially observable environments.

\section{COIN: Chain Of INteraction Benchmark}

In this section, we introduce: the formulation of tasks in COIN (\ref{subsec:task-definination}), how we built such tasks in COIN (\ref{sec:tasks_building}), how we collected datasets with human-in-the-loop teleoperation (\ref{subsec:data-collection}), the statistics of COIN (\ref{subsec:statistics}), and the evaluation metrics (\ref{subsec:metrics}). 

\subsection{Tasks Formulation}
\label{subsec:task-definination} \label{sec:formulation}
We formalize interactive reasoning tasks as a Partially Observable Markov Decision Process (POMDP): $\mathcal{M} = \langle \mathcal{S}, \mathcal{A}, \mathcal{T}, \mathcal{R}, \mathcal{O}, \mathcal{Z} \rangle $. The state space $\mathcal{S}$ encompasses robot configuration, object states and physical properties. We implement two action spaces, the same as ManiSkill3: for VLA models, $A_{\text{VLA}} = \{\Delta p, \Delta R, g\} \in \mathbb{R}^3 \times SO(3) \times \{0,1\}$ using delta end-effector poses; for CodeAsPolicy approaches, $A_{\text{VLM}} = \{q_1...q_7, g\} \in [q_{\text{min}}, q_{\text{max}}]^7 \times \{0,1\}$ using absolute joint positions. The transition function $\mathcal{T}$ models state dynamics while the reward function $\mathcal{R}: \mathcal{S} \times \mathcal{A} \rightarrow \{0, 1\}$ provides sparse binary success feedback. Observations $\mathcal{O}$ include five camera views (front, left/right back, left front, and wrist-mounted) with depth and segmentation maps, language instructions for the task, and robot proprioceptive data, enabling agents to infer occluded state information through interaction. More details can be found in Appendix~\ref{sec:env_setup}.

\subsection{Task Building}\label{sec:tasks_building}
\textbf{COIN comprises 3 categories and 90 total tasks.}~~We design a hierarchical task structure that systematically evaluates interactive reasoning capabilities across different complexity levels:
\begin{itemize}
\item \textbf{COIN-Primitive (20 tasks)}: Fundamental manipulation skills extracted from interactive reasoning tasks by identifying commonly recurring behavioral patterns and essential manipulation primitives (open-close, pick-place, push-pull, rotation).
\item \textbf{COIN-Composition (20 tasks)}: Mid-term complexity tasks that bridge the gap between primitives and full interactive reasoning, introducing controlled increases in complexity through small visual differences or instruction variations.
\item \textbf{COIN-50 (50 tasks)}: Full interactive reasoning tasks requiring multi-step causal reasoning under partial observability, where agents must continually interact with the environment to gather information and adapt their strategies.
\end{itemize}

\begin{figure}[t!]
\includegraphics[width=1\linewidth]{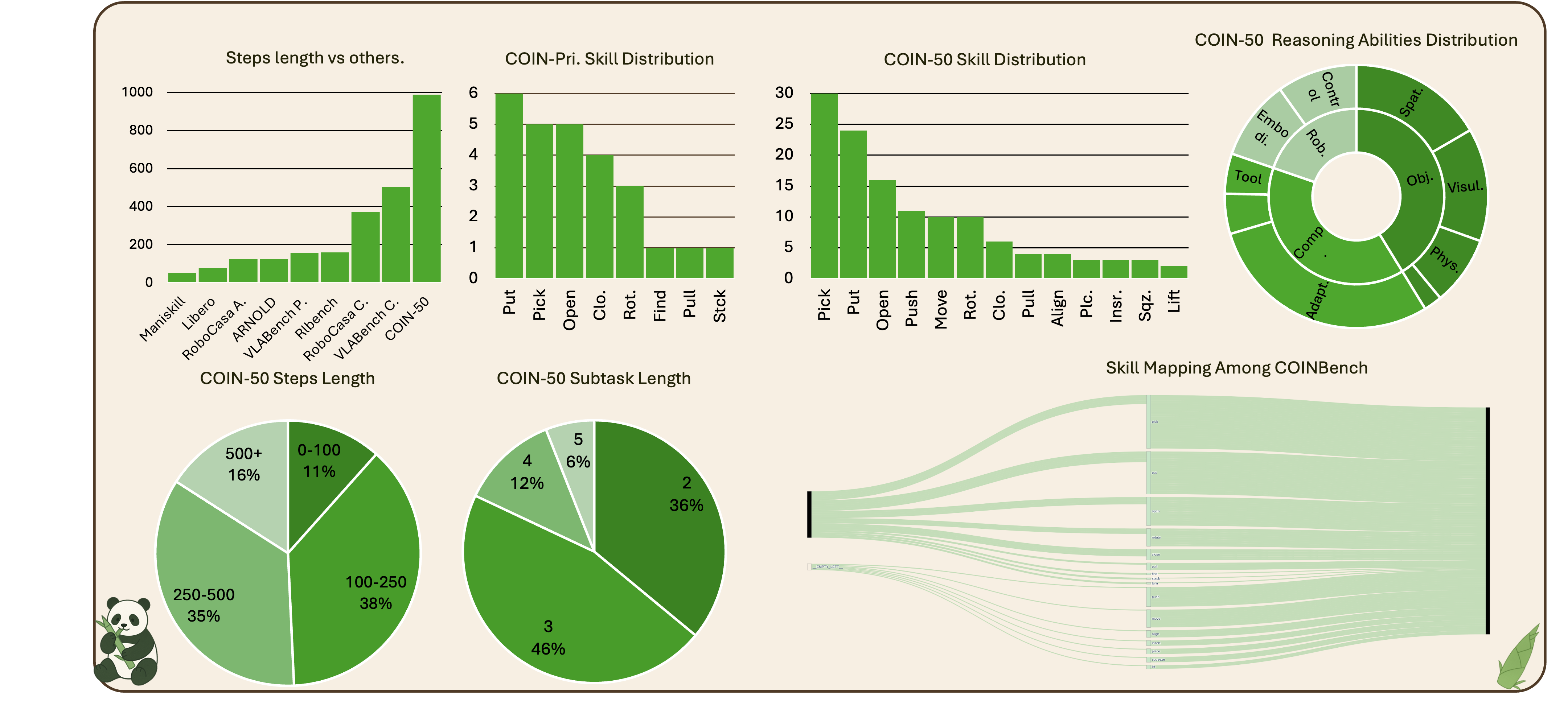}
\caption{Tasks in COIN: we provide diverse tasks with feasible primitive tasks, and provide GT planning for these tasks, which could be used to guide the planning}
\label{fig:taskplot}
\end{figure}

As shown in Figure~\ref{fig:taskplot} and Appendix~\ref{subsec:task-class}, we categorized these interactive tasks into three main domains: (1) object information perception and manipulation, (2) robot-understanding and control, and (3) compositional reasoning. This taxonomy helps systematically evaluate different aspects of interactive reasoning capabilities in embodied agents.

\textbf{Technical implementation.}~~We built all environments on the ManiSkill3 platform~\cite{tao_maniskill3_nodate}, using the Franka Emika Panda robot in tabletop manipulation settings. Environmental assets include articulated objects from PartNet-Mobility~\cite{xiang2020sapien} and additional assets from~\cite{zeng_transporter_2022, li_evaluating_2024, sketchfab}. All 90 tasks provide language instructions and corresponding reward.

\textbf{Subtasks and VQA}~~For each interactive task, we provide: (1) expert demonstrations, (2) ground truth planning in the form of decomposed subtask sequences (\emph{oracle manipulation flow}). As illustrated in Figure~\ref{fig:taskplot}, the average subtask length is 2.83, highlighting the multi-stage nature of these tasks. (3) VQA (Visual Question-Answering) evaluations. The VQA is similar to ERQA protocol~\cite{nvidia2025cosmosreason1physicalcommonsense, erqa2024} queries VLMs with task-specific questions about success conditions or interaction history, serving as an embodied reasoning probe for future research. We formulated them as multiple-choice problems, where VLMs answer these questions by selecting the right answer.

\subsection{Teleoperation and Data Collection}
\label{subsec:data-collection}

\textbf{Low-cost mobile AR teleoperation system.}~~We introduce COIN-teleoperation, a smartphone-based teleoperation system with a total hardware cost under \$20. Built on ARKit~\cite{apple_arkit} and ARCore~\cite{google_arcore} with the help of \cite{ayyan2024mujocoar}, this system captures 6-DoF pose data from mobile devices and achieves stable 20Hz control frequency even on older phones (e.g., iPhone 7 Plus), making robotic data collection broadly accessible. Our comprehensive validation \ref{sec:teleoperation} demonstrates 90\% data replay success and cross-device compatibility, confirming the reliability of our collection approach.

\textbf{COIN-Primitive Dataset.}~~We collected a comprehensive dataset of 20 \textbf{COIN-Primitive} tasks as mentioned in with 50 demonstrations per task captured from 5 camera viewpoints, totaling 1,000 trajectories using COIN-teleoperation. This dataset serves as the primary training resource for VLA model fine-tuning, providing diverse manipulation primitives that form the building blocks for more complex interactive reasoning tasks.
\subsection{COIN statistics}
\vspace{-3mm}
\label{subsec:statistics}

\textbf{Statistics.}~~Figure~\ref{fig:taskplot} presents a comprehensive overview of COIN's benchmark structure. COIN-50 features an average task length of approximately 990 steps, substantially longer than existing benchmarks. More critically, each task requires an average of 2.83 subtasks with frequent interactive reasoning cycles, where 36\% of tasks contain 2 subtasks, 46\% contain 3, and 12\% contain 4. This reveals that our benchmark poses greater challenges not merely through temporal extension, but through the density of reasoning interactions required—necessitating iterative "interaction-reasoning-interaction" loops rather than simple sequential execution, fundamentally distinguishing interactive reasoning from purely long-horizon tasks.

The benchmark's reasoning taxonomy spans object-centric, robot-centric, and compositional reasoning. This focus addresses the under-representation of interactive reasoning in prior benchmarks and supports the modeling of complex "interaction-reasoning-interaction" loops. Overall, COIN offers a comprehensive and realistic testbed for assessing manipulation skills and reasoning capabilities under partial observability and task complexity.

\subsection{Evaluation Metrics}
\label{subsec:metrics}
We introduce a comprehensive evaluation framework with six complementary metrics that assess different aspects of interactive reasoning and manipulation performance across all COIN tasks:

\textbf{Task Performance Metrics:}
\begin{itemize}
\item \textbf{Success Rate (SR)} measures the proportion of successfully completed trajectories across all evaluated tasks.
\item \textbf{Class Success Rate (CSR)} measures category-specific performance across reasoning domains (object-centric, robot-centric, compositional).
\end{itemize}

\textbf{Reasoning Assessment Metrics:}
\begin{itemize}
\item \textbf{Visual Question Answering Score (VS)} assesses perceptual and reasoning capabilities by evaluating whether models correctly answer questions about environmental states and interactive consequences.
\end{itemize}

\textbf{Fine-grained Execution Quality Metrics:}
\begin{itemize}
\item \textbf{Trajectory Stability Score (TS)} measures action quality and smoothness to identify erratic VLA behaviors:
$$TS = 0.3 \cdot S_{vel} + 0.3 \cdot S_{acc} + 0.4 \cdot S_{jerk}$$
where each component uses $\text{Smooth}(x) = \exp(-CV_x)$ with $CV_x = \frac{\sigma_x}{\mu_x}$ (coefficient of variation) applied to velocity, acceleration, jerk (3rd derivative), and position respectively. Higher scores indicate better trajectory stability.

\item \textbf{Gripper Control Stability (GS)} assesses manipulation quality through coordination analysis:
$$GS = 0.4 \cdot S_{smooth} + 0.3 \cdot S_{freq} + 0.3 \cdot S_{coord}$$
where $S_{smooth} = \exp(-\text{abrupt changes})$ penalizes sudden gripper state transitions, $S_{freq} = \exp(-\frac{N_{changes}}{N_{expected}})$ evaluates action frequency appropriateness, and $S_{coord}$ analyzes arm-gripper coordination timing. Higher scores indicate better gripper control quality.

\item \textbf{Generalization Capability Score (GCS)} evaluates model adaptability through controlled task variations:
$$GCS = \frac{\text{SR}_{\text{composition}}}{\text{SR}_{\text{primitive}}}$$
where success rates are averaged across all tasks in each category. Scores close to 1.0 indicate good generalization; lower scores reveal generalization failures.
\end{itemize}

These metrics provide comprehensive evaluation across task completion, reasoning understanding, execution quality, and generalization capability, enabling detailed analysis of model performance across all COIN benchmark components. 

\begin{figure}
\centering
\includegraphics[width=0.7\linewidth]{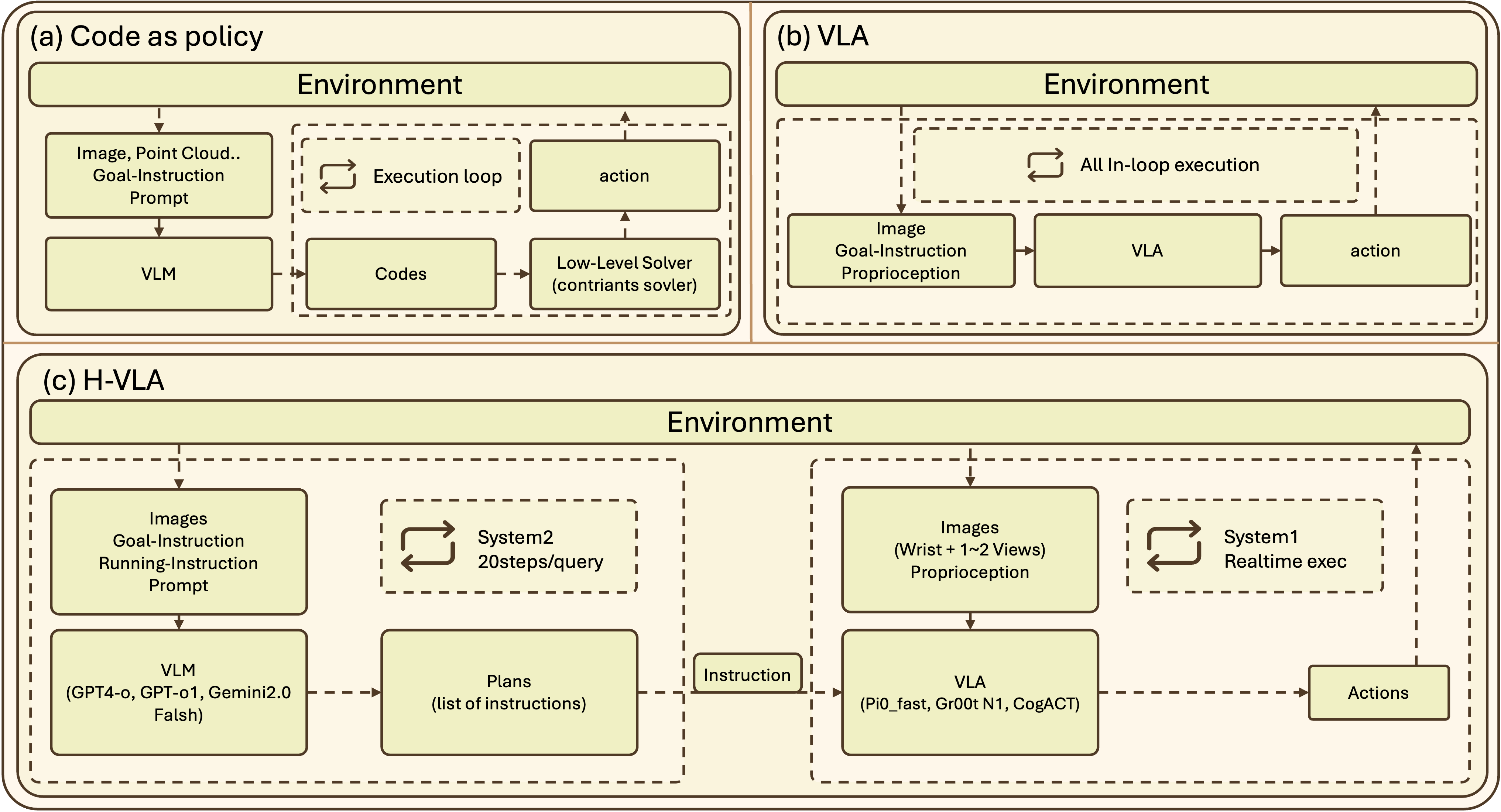}
\caption{Model Architecture Comparison: \textbf{(a)} CodeAsPolicy uses VLMs for planning, with execution handled separately by low-level code and constraint optimizers. \textbf{(b)} End-to-End VLA performs in-loop perception and action directly from the environment. \textbf{(c)} Hierarchical VLA (H-VLA) combines high-level planning (System 2) with low-level VLA execution (System 1), connected via language instructions.}
\label{fig:hvla}
\vspace{-3mm}
\end{figure}

\subsection{Hierarchical VLA (H-VLA) Architecture for COIN}\label{sec:hvla}

Similar to Helix~\cite{helix}, we propose a two-layered VLA framework that decomposes complex reasoning tasks into manageable skill sequences, illustrated in Figure~\ref{fig:hvla}(c).

\begin{itemize}
\item \textbf{System 2 (High-level planner)}, a VLM that processes multi-view images and task instructions to generate a sequence of sub-tasks. Operating at fixed intervals, it monitors execution progress by periodically evaluating current observations and adjusting the instruction queue accordingly.

\item \textbf{System 1 (Low-level executor)}, a VLA model that converts individual skill instructions into robot actions. Taking images, proprioception data, and the current instruction as input, it generates actions in real-time without knowledge of the overall task plan.
\end{itemize}
\section{Experiments}
\label{sec:experiments}

In this section, we evaluate model performance across different task sets. Section~\ref{subsec:exp-setup} introduces our experiment setup and the models tested on COIN (see Figure~\ref{fig:hvla} for an overview). We first analyze how H-VLA and CodeAsPolicy perform on the complex COIN-50 tasks in Section~\ref{sec:coin_50_ana}. Since most models struggle with COIN-50, we then examine their abilities on basic manipulation tasks in COIN-Primitive and COIN-Composition to better understand the causes of failure (Section~\ref{sec:coin_pri_performance}).
\subsection{Experimental Setup}
\label{subsec:exp-setup}

\textbf{Models for COIN-50.}~~COIN-50's complex interactive reasoning tasks require models capable of adaptive planning and execution. We evaluate:
\begin{itemize}
	 \item \textbf{H-VLA models.}~~As described in Section~\ref{sec:hvla}, this two-tier architecture combines VLMs for high-level planning with VLAs for execution. We evaluate six configurations pairing two high-level planners (\textbf{GPT-4o} and \textbf{Gemini 2.0 Flash}) with three VLA models (\textbf{Gr00t N1}, \textbf{Pi0}, and \textbf{CogACT}). Unlike end-to-end VLAs, H-VLA can update plans during execution as new information becomes available through interaction.
	 \item \textbf{CodeAsPolicy approaches.}~~We implement two code-based planning systems: \textbf{Voxposer} and \textbf{Rekep}, both using \texttt{gpt-4o-2024-11-20} for task decomposition and execution planning. Each system reconstructs the environment from three camera views, with Voxposer additionally utilizing ground truth object lists to enhance scene understanding. These approaches separate perception and planning from execution through programmatic interfaces.
\end{itemize}

\textbf{Models for COIN-Primitive and COIN-Composition.}~~We only consider the "low-level controller" of the two families of models above for COIN-50. Effectively, these are end-to-end VLA models as in H-VLA models (see Section~\ref{sec:hvla}) and CodeAsPolicy itself, which is the same model on different benchmarks. 

\begin{itemize}
	 \item \textbf{End-to-end VLA models.}~~We evaluate 3 cutting-edge vision-language-action models as adopted in H-VLA above: \textbf{Gr00t N1}~\cite{nvidia_gr00t_2025}, \textbf{Pi0}~\cite{openpi}, and \textbf{CogACT}~\cite{li_cogact_2024}. Both Gr00t N1 and Pi0 process multi-view observations from three cameras (base-front, left-front, and wrist-mounted), while CogACT processes only the left-front view per its design requirements. 
	 We fine-tune all VLA models on the COIN-Primitive dataset until convergence or for a maximum of three days (see Appendix~\ref{subsec:model-configs} for details). Following \cite{liu_libero_2023}, we select checkpoints based on validation success rates. 

	 \item \textbf{CodeAsPolicy approaches.}~~We evaluate the same Voxposer and Rekep implementations on COIN-Primitive tasks to assess their performance on fundamental manipulation skills.
\end{itemize}

\textbf{Evaluation Details.}~~We report \textbf{SR} averaged over 10 trials. \textbf{CSR} is generated from the \textbf{SR}, and the VQA score is generated by querying the VLM with expert demonstrations for about 50 steps per query. For \textbf{TS} and \textbf{GS}, we report the scores according to the recorded trajectories during evaluation. For \textbf{GCS}, we evaluate the score according to the \textbf{SR} between \textbf{COIN-Primitive} and \textbf{COIN-Composition}. All tasks and environment specifications can be found in Appendix~\ref{sec:env_setup} and \ref{subsec:task-class}. 
 
\subsection{Main results for COIN-50}
\label{sec:coin_50_ana}

\begin{table}[t]
\caption{Trajectory and gripper stability analysis across different task types. Values show mean $\pm$ standard deviation, revealing execution quality patterns across model architectures. \textbf{Bold values} indicate performance exceeding human baseline.}
\label{tab:stability-analysis}
\centering
\small
\begin{tabular}{lccc}
\toprule
\textbf{Model} & \textbf{Task Type} & \textbf{Trajectory Stability} & \textbf{Gripper Stability} \\
\midrule
CogACT & Primitive & $\mathbf{0.150 \pm 0.055}$ & $\mathbf{0.872 \pm 0.134}$ \\
CogACT & Composition & $\mathbf{0.138 \pm 0.039}$ & $\mathbf{0.796 \pm 0.136}$ \\
CogACT & Interactive & $\mathbf{0.146 \pm 0.041}$ & $\mathbf{0.782 \pm 0.141}$ \\
\midrule
Gr00t N1 & Primitive & $0.082 \pm 0.015$ & $0.318 \pm 0.116$ \\
Gr00t N1 & Composition & $0.086 \pm 0.002$ & $0.327 \pm 0.058$ \\
Gr00t N1 & Interactive & $0.084 \pm 0.002$ & $0.294 \pm 0.050$ \\
\midrule
Pi0 & Primitive & $0.084 \pm 0.067$ & $0.440 \pm 0.198$ \\
Pi0 & Composition & $0.035 \pm 0.043$ & $0.465 \pm 0.253$ \\
Pi0 & Interactive & $0.061 \pm 0.050$ & $0.440 \pm 0.219$ \\
\midrule
Human Dataset & Primitive & $0.134 \pm 0.035$ & $0.684 \pm 0.297$ \\
\bottomrule
\end{tabular}
\vspace{-3mm}
\end{table}

\textbf{Overview: Interactive reasoning remains a fundamental challenge for current AI approaches.} Our evaluation reveals a stark capability gap in all tested systems when faced with tasks requiring interactive reasoning. As shown in Table~\ref{tab:interactive-success}, all models fail to solve complex interactive reasoning tasks, with success rates rarely exceeding 3\%. Our analysis on COIN-50 reveals fundamental limitations in both major approach categories:

\label{exp:code_as_policy}
\textbf{CodeAsPolicy approaches face two critical issues:} (1) \textit{Non-interactive planning architecture}: These methods cannot update plans based on environmental feedback, making them fundamentally unsuited for partially observable environments requiring iterative interaction. For example, in "Pick the cube" task, if the cube was not picked, it only repeat "back-home" and "pick the cube" loop, without any new strategies. (2) \textit{Planning-execution gap}: Significant disconnects exist as shown in \cite{huang_rekep_2024} between high-level plans and low-level execution capabilities, which we analyze in detail through COIN-Primitive and COIN-Composition evaluations.

\textbf{H-VLA approaches suffer from multiple limitations:} (1) \textit{Poor VLM planning performance}: High-level reasoning and plan generation capabilities are insufficient for complex interactive scenarios, and their performance does not improve significantly as the number of interaction steps increases. (2) \textit{Inadequate VLA execution}: Low-level action generation models demonstrate poor manipulation capabilities. (3) \textit{Weak VLM-VLA integration}: The coordination between high-level planning and low-level execution remains problematic. The integration relies on natural language instructions solely, which is not adequate to represent the complex interaction between the robot and the environment as shown in \cite{instruction_consistency}. Points 2 and 3 will be discussed in the next section.

To understand these fundamental limitations, we conducted detailed evaluations on COIN-Primitive and COIN-Composition tasks. The following analysis provides deeper insights into the specific failure modes of each approach.

\subsection{Main Results on COIN-Primitive and COIN-Composition}
\label{sec:coin_pri_performance}
\label{sec:coin-pri}

\textbf{Overview: COIN-Primitive and COIN-Composition reveal specific failure modes.} COIN-Primitive serves as a testbed for evaluating fundamental manipulation skills, while COIN-Composition tests generalization to minor environmental variations. Our detailed analysis confirms the limitations identified in COIN-50 and reveals specific failure modes for each approach category.

\begin{figure}
\centering
\includegraphics[width=0.85\linewidth]{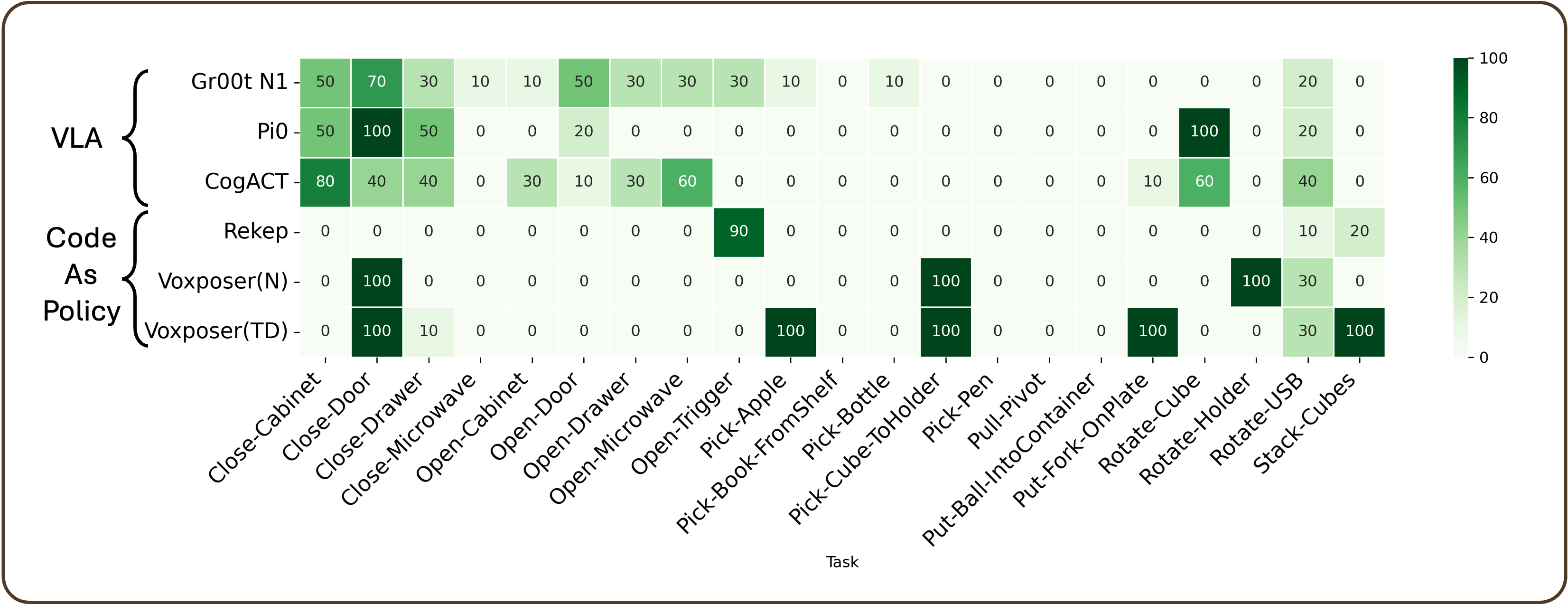}
\caption{Performance heatmap for different models on COIN-Primitive tasks. The visualization reveals that VLA models achieve broader task coverage than CodeAsPolicy approaches, though with different strengths across task types. Color intensity indicates success rate.}
\label{fig:heatmap}
\vspace{-3mm}
\end{figure}

\textbf{CodeAsPolicy approaches reveal planning-execution disconnects.} Our evaluation confirms the two critical issues identified in COIN-50: (1) \textit{VLM-executor gap}: Significant disconnects exist between high-level VLM planning and low-level execution capabilities. As shown in Figure~\ref{fig:heatmap}, Voxposer and Rekep perform poorly even on basic manipulation tasks, indicating fundamental misalignment between planning and execution.  
(2) Articuation Manipulation problems: These two models are not able to handle the articulated objects, such as cabinet, doors and switchs. This is mainly caused by the structure is not feasible for key-points based representation.

\textbf{H-VLA approaches confirm the three limitations identified in COIN-50.} Our detailed analysis reveals specific manifestations of the issues identified earlier:
(1) \textit{Poor trajectory and gripper control}: VLAs exhibit severe control precision issues, particularly struggling with gripper timing~\cite{nvidia_gr00t_2025}. As shown in Table~\ref{tab:stability-analysis}, CogACT demonstrates relatively stable trajectories compared to other VLAs, potentially due to its temporal ensemble mechanism, which is left for discussion in future work. However, all VLAs show significant jerky movements and high discontinuity.
(2) \textit{Catastrophic generalization failures}: As shown in Table~\ref{tab:generalization}, models achieving reasonable success on primitive tasks (16-19\%) experience complete failure when faced with composition tasks. Even adding a single new object or switching instructions causes task failure. 
(3) \textit{Weak VLM-VLA integration}: Despite broader task coverage when overfitted to primitive tasks, the coordination between high-level planning and low-level execution remains fundamentally problematic. For example, while VLAs can successfully execute "open the door" commands, changing the instruction to "pull the door" for the same physical action results in dramatically reduced success rates and moves the gripper to unreasonable locations, demonstrating that the natural language interface fails to capture the underlying action semantics.
\begin{table}[t]
\caption{Generalization capability evaluation using COIN-Composition tasks. Models demonstrate severe generalization failures when faced with minor visual or instruction variations from primitive tasks.}
\label{tab:generalization}
\centering
\begin{tabular}{lcccc}
\toprule
\textbf{Model} & \textbf{Primitive SR} & \textbf{Composition SR} & \textbf{Finished Tasks} & \textbf{GCS} \\
\midrule
CogACT & 19.0\% & 1.5\% & 3/20 & 0.079 \\
Pi0 & 16.1\% & 6.5\% & 4/20 & 0.404 \\
Gr00t N1 & 16.7\% & 0.0\% & 0/20 & 0.000 \\
\bottomrule
\end{tabular}
\vspace{-3mm}
\end{table}

\section{Conclusions}
We introduce COIN, an evaluation benchmark for interactive reasoning in embodied AI with three levels: COIN-Primitive, COIN-Composition, and COIN-50. COIN also includes a teleoperation pipeline and 1,000 demonstration trajectories for model training.

Our results show that interactive reasoning remains a major bottleneck for current embodied systems. CodeAsPolicy methods struggle to adapt through interaction, while H-VLA systems remain limited by unstable control, weak generalization, and poor coordination between planning and execution.

These findings suggest four immediate directions for improvement: smoother VLA control, stronger multimodal perception, tighter VLM-VLA integration beyond language-only interfaces, and adaptive CodeAsPolicy frameworks with closed-loop feedback.
\newpage
\bibliographystyle{iclr2026_conference}
\bibliography{new_ref_dedup}

@Misc{		  brohan_rt-1_2023,
  title		= {{RT}-1: Robotics Transformer for Real-World Control at
		  Scale},
  url		= {http://arxiv.org/abs/2212.06817},
  doi		= {10.48550/arXiv.2212.06817},
  shorttitle	= {{RT}-1},
  number	= {{arXiv}:2212.06817},
  publisher	= {{arXiv}},
  author	= {Brohan, Anthony and Brown, Noah and Carbajal, Justice and
		  Chebotar, Yevgen and Dabis, Joseph and Finn, Chelsea and
		  Gopalakrishnan, Keerthana and Hausman, Karol and Herzog,
		  Alex and Hsu, Jasmine and Ibarz, Julian and Ichter, Brian
		  and Irpan, Alex and Jackson, Tomas and Jesmonth, Sally and
		  Joshi, Nikhil J. and Julian, Ryan and Kalashnikov, Dmitry
		  and Kuang, Yuheng and Leal, Isabel and Lee, Kuang-Huei and
		  Levine, Sergey and Lu, Yao and Malla, Utsav and Manjunath,
		  Deeksha and Mordatch, Igor and Nachum, Ofir and Parada,
		  Carolina and Peralta, Jodilyn and Perez, Emily and Pertsch,
		  Karl and Quiambao, Jornell and Rao, Kanishka and Ryoo,
		  Michael and Salazar, Grecia and Sanketi, Pannag and Sayed,
		  Kevin and Singh, Jaspiar and Sontakke, Sumedh and Stone,
		  Austin and Tan, Clayton and Tran, Huong and Vanhoucke,
		  Vincent and Vega, Steve and Vuong, Quan and Xia, Fei and
		  Xiao, Ted and Xu, Peng and Xu, Sichun and Yu, Tianhe and
		  Zitkovich, Brianna},
  urldate	= {2025-02-25},
  date		= {2023-08-11},
  eprinttype	= {arxiv},
  eprint	= {2212.06817 [cs]}
}

@Misc{		  rt2,
  title		= {{RT}-2: Vision-Language-Action Models Transfer Web
		  Knowledge to Robotic Control},
  url		= {http://arxiv.org/abs/2307.15818},
  doi		= {10.48550/arXiv.2307.15818},
  shorttitle	= {{RT}-2},
  number	= {{arXiv}:2307.15818},
  publisher	= {{arXiv}},
  author	= {Brohan, Anthony and Brown, Noah and Carbajal, Justice and
		  Chebotar, Yevgen and Chen, Xi and Choromanski, Krzysztof
		  and Ding, Tianli and Driess, Danny and Dubey, Avinava and
		  Finn, Chelsea and Florence, Pete and Fu, Chuyuan and
		  Arenas, Montse Gonzalez and Gopalakrishnan, Keerthana and
		  Han, Kehang and Hausman, Karol and Herzog, Alexander and
		  Hsu, Jasmine and Ichter, Brian and Irpan, Alex and Joshi,
		  Nikhil and Julian, Ryan and Kalashnikov, Dmitry and Kuang,
		  Yuheng and Leal, Isabel and Lee, Lisa and Lee, Tsang-Wei
		  Edward and Levine, Sergey and Lu, Yao and Michalewski,
		  Henryk and Mordatch, Igor and Pertsch, Karl and Rao,
		  Kanishka and Reymann, Krista and Ryoo, Michael and Salazar,
		  Grecia and Sanketi, Pannag and Sermanet, Pierre and Singh,
		  Jaspiar and Singh, Anikait and Soricut, Radu and Tran,
		  Huong and Vanhoucke, Vincent and Vuong, Quan and Wahid,
		  Ayzaan and Welker, Stefan and Wohlhart, Paul and Wu, Jialin
		  and Xia, Fei and Xiao, Ted and Xu, Peng and Xu, Sichun and
		  Yu, Tianhe and Zitkovich, Brianna},
  urldate	= {2025-02-11},
  date		= {2023-07-28},
  langid	= {english},
  eprinttype	= {arxiv},
  eprint	= {2307.15818 [cs]}
}

@Misc{		  duan_manipulate-anything_2024,
  title		= {Manipulate-Anything: Automating Real-World Robots using
		  Vision-Language Models},
  url		= {http://arxiv.org/abs/2406.18915},
  doi		= {10.48550/arXiv.2406.18915},
  shorttitle	= {Manipulate-Anything},
  number	= {{arXiv}:2406.18915},
  publisher	= {{arXiv}},
  author	= {Duan, Jiafei and Yuan, Wentao and Pumacay, Wilbert and
		  Wang, Yi Ru and Ehsani, Kiana and Fox, Dieter and Krishna,
		  Ranjay},
  urldate	= {2025-02-14},
  date		= {2024-08-29},
  eprinttype	= {arxiv},
  eprint	= {2406.18915 [cs]}
}

@Misc{		  feng_reflective_2025,
  title		= {Reflective Planning: Vision-Language Models for
		  Multi-Stage Long-Horizon Robotic Manipulation},
  url		= {http://arxiv.org/abs/2502.16707},
  doi		= {10.48550/arXiv.2502.16707},
  shorttitle	= {Reflective Planning},
  number	= {{arXiv}:2502.16707},
  publisher	= {{arXiv}},
  author	= {Feng, Yunhai and Han, Jiaming and Yang, Zhuoran and Yue,
		  Xiangyu and Levine, Sergey and Luo, Jianlan},
  urldate	= {2025-02-26},
  date		= {2025-02-23},
  eprinttype	= {arxiv},
  eprint	= {2502.16707 [cs]}
}

@InProceedings{	  haresh_clevrskills_2024,
  title		= {{ClevrSkills}: Compositional Language And Visual Reasoning
		  in Robotics},
  url		= {https://openreview.net/forum?id=64sZtFSOh6#discussion},
  shorttitle	= {{ClevrSkills}},
  eventtitle	= {The Thirty-eight Conference on Neural Information
		  Processing Systems Datasets and Benchmarks Track},
  author	= {Haresh, Sanjay and Dijkman, Daniel and Bhattacharyya,
		  Apratim and Memisevic, Roland},
  urldate	= {2025-02-14},
  date		= {2024-11-13},
  langid	= {english}
}

@Misc{		  copa,
  title		= {{CoPa}: General Robotic Manipulation through Spatial
		  Constraints of Parts with Foundation Models},
  url		= {http://arxiv.org/abs/2403.08248},
  doi		= {10.48550/arXiv.2403.08248},
  shorttitle	= {{CoPa}},
  number	= {{arXiv}:2403.08248},
  publisher	= {{arXiv}},
  author	= {Huang, Haoxu and Lin, Fanqi and Hu, Yingdong and Wang,
		  Shengjie and Gao, Yang},
  urldate	= {2025-02-11},
  date		= {2024-03-13},
  langid	= {english},
  eprinttype	= {arxiv},
  eprint	= {2403.08248 [cs]}
}

@Misc{		  huang_rekep_2024,
  title		= {{ReKep}: Spatio-Temporal Reasoning of Relational Keypoint
		  Constraints for Robotic Manipulation},
  url		= {http://arxiv.org/abs/2409.01652},
  doi		= {10.48550/arXiv.2409.01652},
  shorttitle	= {{ReKep}},
  number	= {{arXiv}:2409.01652},
  publisher	= {{arXiv}},
  author	= {Huang, Wenlong and Wang, Chen and Li, Yunzhu and Zhang,
		  Ruohan and Fei-Fei, Li},
  urldate	= {2025-02-11},
  date		= {2024-11-12},
  langid	= {english},
  eprinttype	= {arxiv},
  eprint	= {2409.01652 [cs]}
}

@Misc{		  voxposer,
  title		= {{VoxPoser}: Composable 3D Value Maps for Robotic
		  Manipulation with Language Models},
  url		= {http://arxiv.org/abs/2307.05973},
  doi		= {10.48550/arXiv.2307.05973},
  shorttitle	= {{VoxPoser}},
  number	= {{arXiv}:2307.05973},
  publisher	= {{arXiv}},
  author	= {Huang, Wenlong and Wang, Chen and Zhang, Ruohan and Li,
		  Yunzhu and Wu, Jiajun and Fei-Fei, Li},
  urldate	= {2025-02-11},
  date		= {2023-11-02},
  langid	= {english},
  eprinttype	= {arxiv},
  eprint	= {2307.05973 [cs]}
}

@Misc{		  li_cogact_2024,
  title		= {{CogACT}: A Foundational Vision-Language-Action Model for
		  Synergizing Cognition and Action in Robotic Manipulation},
  url		= {http://arxiv.org/abs/2411.19650},
  doi		= {10.48550/arXiv.2411.19650},
  shorttitle	= {{CogACT}},
  number	= {{arXiv}:2411.19650},
  publisher	= {{arXiv}},
  author	= {Li, Qixiu and Liang, Yaobo and Wang, Zeyu and Luo, Lin and
		  Chen, Xi and Liao, Mozheng and Wei, Fangyun and Deng, Yu
		  and Xu, Sicheng and Zhang, Yizhong and Wang, Xiaofan and
		  Liu, Bei and Fu, Jianlong and Bao, Jianmin and Chen, Dong
		  and Shi, Yuanchun and Yang, Jiaolong and Guo, Baining},
  urldate	= {2025-02-27},
  date		= {2024-11-29},
  eprinttype	= {arxiv},
  eprint	= {2411.19650 [cs]}
}

@Misc{		  li_evaluating_2024,
  title		= {Evaluating Real-World Robot Manipulation Policies in
		  Simulation},
  url		= {http://arxiv.org/abs/2405.05941},
  doi		= {10.48550/arXiv.2405.05941},
  number	= {{arXiv}:2405.05941},
  publisher	= {{arXiv}},
  author	= {Li, Xuanlin and Hsu, Kyle and Gu, Jiayuan and Pertsch,
		  Karl and Mees, Oier and Walke, Homer Rich and Fu, Chuyuan
		  and Lunawat, Ishikaa and Sieh, Isabel and Kirmani, Sean and
		  Levine, Sergey and Wu, Jiajun and Finn, Chelsea and Su, Hao
		  and Vuong, Quan and Xiao, Ted},
  urldate	= {2025-02-21},
  date		= {2024-05-09},
  langid	= {english},
  eprinttype	= {arxiv},
  eprint	= {2405.05941 [cs]}
}

@Misc{		  liu_libero_2023,
  title		= {{LIBERO}: Benchmarking Knowledge Transfer for Lifelong
		  Robot Learning},
  url		= {http://arxiv.org/abs/2306.03310},
  doi		= {10.48550/arXiv.2306.03310},
  shorttitle	= {{LIBERO}},
  number	= {{arXiv}:2306.03310},
  publisher	= {{arXiv}},
  author	= {Liu, Bo and Zhu, Yifeng and Gao, Chongkai and Feng, Yihao
		  and Liu, Qiang and Zhu, Yuke and Stone, Peter},
  urldate	= {2025-03-08},
  date		= {2023-10-14},
  langid	= {english},
  eprinttype	= {arxiv},
  eprint	= {2306.03310 [cs]}
}

@Misc{		  mees_calvin_2022,
  title		= {{CALVIN}: A Benchmark for Language-Conditioned Policy
		  Learning for Long-Horizon Robot Manipulation Tasks},
  url		= {http://arxiv.org/abs/2112.03227},
  doi		= {10.48550/arXiv.2112.03227},
  shorttitle	= {{CALVIN}},
  number	= {{arXiv}:2112.03227},
  publisher	= {{arXiv}},
  author	= {Mees, Oier and Hermann, Lukas and Rosete-Beas, Erick and
		  Burgard, Wolfram},
  urldate	= {2025-03-08},
  date		= {2022-07-13},
  langid	= {english},
  eprinttype	= {arxiv},
  eprint	= {2112.03227 [cs]}
}

@Article{	  tao_maniskill3_nodate,
  title		= {{MANISKILL}3: {GPU} {PARALLELIZED} {ROBOTICS} {SIMULATION}
		  {AND} {RENDERING} {FOR} {GENERALIZABLE} {EMBODIED} {AI}},
  author	= {Tao, Stone and Xiang, Fanbo and Shukla, Arth and Qin,
		  Yuzhe and Hinrichsen, Xander and Yuan, Xiaodi and Lin}}

@Article{li2024evaluating,
  title = {Evaluating Real-World Robot Manipulation Policies in Simulation},
  author = {Li, Chengshu and others},
  year = {2024},
  url = {https://arxiv.org/abs/2405.05941},
  eprint = {2405.05941},
  archivePrefix = {arXiv}
}

@Article{code_as_policy,
  title = {Code as Policies: Language Model Programs for Embodied Control},
  author = {Liang, Percy and Chen, Rishi and Huang, Po-Hsun and Vatsal, Nikhil and Xu, Pieter and Zhou, Tete and Zhou, Yuhui and Xia, Fei and Hausman, Karol and Ichter, Brian and others},
  year = {2022},
  url = {https://arxiv.org/abs/2209.07753},
  eprint = {2209.07753},
  archivePrefix = {arXiv}
}

@Article{sam2,
  title = {Segment Anything Model 2},
  author = {Kirillov, Alexander and Mintun, Eric and Ravi, Nikhila and Mao, Hanzi and Rolland, Chloe and Gustafson, Laura and Xiao, Tete and Whitehead, Spencer and Berg, Alexander C. and Lo, Wan-Yen and others},
  year = {2024},
  url = {https://arxiv.org/abs/2404.14192},
  eprint = {2404.14192},
  archivePrefix = {arXiv}
}

@Article{sam,
  title = {Segment Anything},
  author = {Kirillov, Alexander and Mintun, Eric and Ravi, Nikhila and Mao, Hanzi and Rolland, Chloe and Gustafson, Laura and Xiao, Tete and Whitehead, Spencer and Berg, Alexander C. and Lo, Wan-Yen and others},
  year = {2023},
  url = {https://arxiv.org/abs/2304.02643},
  eprint = {2304.02643},
  archivePrefix = {arXiv}
}

@Article{apple_arkit,
  title = {ARKit - Apple Developer},
  author = {Apple Inc.},
  year = {2023},
  url = {https://developer.apple.com/augmented-reality/arkit/},
  note = {Augmented reality framework for iOS}
}

@Article{google_arcore,
  title = {ARCore - Google Developers},
  author = {Google LLC},
  year = {2023},
  url = {https://developers.google.com/ar},
  note = {Augmented reality platform for Android}
}

@Article{		  yang_embodiedbench_2025,
  title		= {{EmbodiedBench}: Comprehensive Benchmarking Multi-modal
		  Large Language Models for Vision-Driven Embodied Agents},
  url		= {http://arxiv.org/abs/2502.09560},
  doi		= {10.48550/arXiv.2502.09560},
  shorttitle	= {{EmbodiedBench}},
  number	= {{arXiv}:2502.09560},
  publisher	= {{arXiv}},
  author	= {Yang, Rui and Chen, Hanyang and Zhang, Junyu and Zhao,
		  Mark and Qian, Cheng and Wang, Kangrui and Wang, Qineng and
		  Koripella, Teja Venkat and Movahedi, Marziyeh and Li,
		  Manling and Ji, Heng and Zhang, Huan and Zhang, Tong},
  urldate	= {2025-02-15},
  date		= {2025-02-13},
  eprinttype	= {arxiv},
  eprint	= {2502.09560 [cs]}
}

@Article{		  zeng_transporter_2022,
  title		= {Transporter Networks: Rearranging the Visual World for
		  Robotic Manipulation},
  url		= {http://arxiv.org/abs/2010.14406},
  doi		= {10.48550/arXiv.2010.14406},
  shorttitle	= {Transporter Networks},
  number	= {{arXiv}:2010.14406},
  publisher	= {{arXiv}},
  author	= {Zeng, Andy and Florence, Pete and Tompson, Jonathan and
		  Welker, Stefan and Chien, Jonathan and Attarian, Maria and
		  Armstrong, Travis and Krasin, Ivan and Duong, Dan and
		  Wahid, Ayzaan and Sindhwani, Vikas and Lee, Johnny},
  urldate	= {2025-02-11},
  date		= {2022-01-05},
  langid	= {english},
  eprinttype	= {arxiv},
  eprint	= {2010.14406 [cs]}
}

@Article{	  zhang_vlabench_nodate,
  title		= {{VLABench}: A Large-Scale Benchmark for
		  Language-Conditioned Robotics Manipulation with
		  Long-Horizon Reasoning Tasks},
  author	= {Zhang, Shiduo and Xu, Zhe and Liu, Peiju and Yu, Xiaopeng
		  and Li, Yuan and Gao, Qinghui and Fei, Zhaoye and Yin,
		  Zhangyue and Wu, Zuxuan and Jiang, Yu-Gang and Qiu,
		  Xipeng},
  langid	= {english}
}

@Article{		  zheng_robocas_2024,
  title		= {{RoboCAS}: A Benchmark for Robotic Manipulation in Complex
		  Object Arrangement Scenarios},
  url		= {http://arxiv.org/abs/2407.06951},
  doi		= {10.48550/arXiv.2407.06951},
  shorttitle	= {{RoboCAS}},
    number	= {{arXiv}:2407.06951},
  publisher	= {{arXiv}},
  author	= {Zheng, Liming and Yan, Feng and Liu, Fanfan and Feng,
		  Chengjian and Kang, Zhuoliang and Ma, Lin},
  urldate	= {2025-02-14},
  date		= {2024-07-09},
  eprinttype	= {arxiv},
  eprint	= {2407.06951 [cs]},
}

@Article{	  zhu_robosuite_nodate,
  title		= {robosuite: A Modular Simulation Framework and Benchmark
		  for Robot Learning},
  author	= {Zhu, Yuke and Wong, Josiah and Mandlekar, Ajay and
		  Martın-Martın, Roberto and Joshi, Abhishek and Nasiriany,
		  Soroush and Zhu, Yifeng},
  langid	= {english}
}

@software{ayyan2024mujocoar,
  author = {Rayyan, Omar},
  title = {{MuJoCo AR: Phone Teleoperation for Robots}},
  url = {https://github.com/omarrayyann/mujocoar},
  version = {1.3.0},
  year = {2024},
}

@misc{nvidia2025cosmosreason1physicalcommonsense,
	  title={Cosmos-Reason1: From Physical Common Sense To Embodied Reasoning}, 
	  author={NVIDIA and : and Alisson Azzolini and Hannah Brandon and Prithvijit Chattopadhyay and Huayu Chen and Jinju Chu and Yin Cui and Jenna Diamond and Yifan Ding and Francesco Ferroni and Rama Govindaraju and Jinwei Gu and Siddharth Gururani and Imad El Hanafi and Zekun Hao and Jacob Huffman and Jingyi Jin and Brendan Johnson and Rizwan Khan and George Kurian and Elena Lantz and Nayeon Lee and Zhaoshuo Li and Xuan Li and Tsung-Yi Lin and Yen-Chen Lin and Ming-Yu Liu and Alice Luo and Andrew Mathau and Yun Ni and Lindsey Pavao and Wei Ping and David W. Romero and Misha Smelyanskiy and Shuran Song and Lyne Tchapmi and Andrew Z. Wang and Boxin Wang and Haoxiang Wang and Fangyin Wei and Jiashu Xu and Yao Xu and Xiaodong Yang and Zhuolin Yang and Xiaohui Zeng and Zhe Zhang},
	  year={2025},
	  eprint={2503.15558},
	  archivePrefix={arXiv},
	  primaryClass={cs.AI},
	  url={https://arxiv.org/abs/2503.15558}, 
}

@article{openpi,
  title = {{$\Pi$}0: {{A Vision-Language-Action Flow Model}} for {{General Robot Control}}},
  author = {Black, Kevin and Brown, Noah and Driess, Danny and Esmail, Adnan and Equi, Michael and Finn, Chelsea and Fusai, Niccolo and Groom, Lachy and Hausman, Karol and Ichter, Brian and Jakubczak, Szymon and Jones, Tim and Ke, Liyiming and Levine, Sergey and {Li-Bell}, Adrian and Mothukuri, Mohith and Nair, Suraj and Pertsch, Karl and Shi, Lucy Xiaoyang and Tanner, James and Vuong, Quan and Walling, Anna and Wang, Haohuan and Zhilinsky, Ury},
  langid = {english}
}

@misc{helix,
  author       = {{Figure AI}},
  title        = {Helix: A Vision-Language-Action Model for Generalist Humanoid Control},
  howpublished = {\url{https://www.figure.ai/news/helix}},
  year         = {2025},
  month        = {February},
  note         = {Accessed: 2025-04-27}
}

@inproceedings{OXE,
  title={Open x-embodiment: Robotic learning datasets and rt-x models: Open x-embodiment collaboration 0},
  author={O’Neill, Abby and Rehman, Abdul and Maddukuri, Abhiram and Gupta, Abhishek and Padalkar, Abhishek and Lee, Abraham and Pooley, Acorn and Gupta, Agrim and Mandlekar, Ajay and Jain, Ajinkya and others},
  booktitle={2024 IEEE International Conference on Robotics and Automation (ICRA)},
  pages={6892--6903},
  year={2024},
  organization={IEEE}
}

@article{droid,
  title={Droid: A large-scale in-the-wild robot manipulation dataset},
  author={Khazatsky, Alexander and Pertsch, Karl and Nair, Suraj and Balakrishna, Ashwin and Dasari, Sudeep and Karamcheti, Siddharth and Nasiriany, Soroush and Srirama, Mohan Kumar and Chen, Lawrence Yunliang and Ellis, Kirsty and others},
  journal={arXiv preprint arXiv:2403.12945},
  year={2024}
}

@misc{geminiroboticsteam2025geminiroboticsbringingai,
	  title={Gemini Robotics: Bringing AI into the Physical World}, 
	  author={Gemini Robotics Team and Saminda Abeyruwan and Joshua Ainslie and Jean-Baptiste Alayrac and Montserrat Gonzalez Arenas and Travis Armstrong and Ashwin Balakrishna and Robert Baruch and Maria Bauza and Michiel Blokzijl and Steven Bohez and Konstantinos Bousmalis and Anthony Brohan and Thomas Buschmann and Arunkumar Byravan and Serkan Cabi and Ken Caluwaerts and Federico Casarini and Oscar Chang and Jose Enrique Chen and Xi Chen and Hao-Tien Lewis Chiang and Krzysztof Choromanski and David D'Ambrosio and Sudeep Dasari and Todor Davchev and Coline Devin and Norman Di Palo and Tianli Ding and Adil Dostmohamed and Danny Driess and Yilun Du and Debidatta Dwibedi and Michael Elabd and Claudio Fantacci and Cody Fong and Erik Frey and Chuyuan Fu and Marissa Giustina and Keerthana Gopalakrishnan and Laura Graesser and Leonard Hasenclever and Nicolas Heess and Brandon Hernaez and Alexander Herzog and R. Alex Hofer and Jan Humplik and Atil Iscen and Mithun George Jacob and Deepali Jain and Ryan Julian and Dmitry Kalashnikov and M. Emre Karagozler and Stefani Karp and Chase Kew and Jerad Kirkland and Sean Kirmani and Yuheng Kuang and Thomas Lampe and Antoine Laurens and Isabel Leal and Alex X. Lee and Tsang-Wei Edward Lee and Jacky Liang and Yixin Lin and Sharath Maddineni and Anirudha Majumdar and Assaf Hurwitz Michaely and Robert Moreno and Michael Neunert and Francesco Nori and Carolina Parada and Emilio Parisotto and Peter Pastor and Acorn Pooley and Kanishka Rao and Krista Reymann and Dorsa Sadigh and Stefano Saliceti and Pannag Sanketi and Pierre Sermanet and Dhruv Shah and Mohit Sharma and Kathryn Shea and Charles Shu and Vikas Sindhwani and Sumeet Singh and Radu Soricut and Jost Tobias Springenberg and Rachel Sterneck and Razvan Surdulescu and Jie Tan and Jonathan Tompson and Vincent Vanhoucke and Jake Varley and Grace Vesom and Giulia Vezzani and Oriol Vinyals and Ayzaan Wahid and Stefan Welker and Paul Wohlhart and Fei Xia and Ted Xiao and Annie Xie and Jinyu Xie and Peng Xu and Sichun Xu and Ying Xu and Zhuo Xu and Yuxiang Yang and Rui Yao and Sergey Yaroshenko and Wenhao Yu and Wentao Yuan and Jingwei Zhang and Tingnan Zhang and Allan Zhou and Yuxiang Zhou},
	  year={2025},
	  eprint={2503.20020},
	  archivePrefix={arXiv},
	  primaryClass={cs.RO},
	  url={https://arxiv.org/abs/2503.20020}, 
}

@misc{nvidia_gr00t_2025,
  title = {{{GR00T N1}}: {{An Open Foundation Model}} for {{Generalist Humanoid Robots}}},
  shorttitle = {{{GR00T N1}}},
  author = {NVIDIA and Bjorck, Johan and Casta{\~n}eda, Fernando and Cherniadev, Nikita and Da, Xingye and Ding, Runyu and Fan, Linxi "Jim" and Fang, Yu and Fox, Dieter and Hu, Fengyuan and Huang, Spencer and Jang, Joel and Jiang, Zhenyu and Kautz, Jan and Kundalia, Kaushil and Lao, Lawrence and Li, Zhiqi and Lin, Zongyu and Lin, Kevin and Liu, Guilin and Llontop, Edith and Magne, Loic and Mandlekar, Ajay and Narayan, Avnish and Nasiriany, Soroush and Reed, Scott and Tan, You Liang and Wang, Guanzhi and Wang, Zu and Wang, Jing and Wang, Qi and Xiang, Jiannan and Xie, Yuqi and Xu, Yinzhen and Xu, Zhenjia and Ye, Seonghyeon and Yu, Zhiding and Zhang, Ao and Zhang, Hao and Zhao, Yizhou and Zheng, Ruijie and Zhu, Yuke},
  year = {2025},
  month = mar,
  number = {arXiv:2503.14734},
  eprint = {2503.14734},
  primaryclass = {cs},
  publisher = {arXiv},
  doi = {10.48550/arXiv.2503.14734},
  urldate = {2025-04-11},
  archiveprefix = {arXiv},
  langid = {english}
}

@article{som,
  title={Set-of-mark prompting unleashes extraordinary visual grounding in gpt-4v},
  author={Yang, Jianwei and Zhang, Hao and Li, Feng and Zou, Xueyan and Li, Chunyuan and Gao, Jianfeng},
  journal={arXiv preprint arXiv:2310.11441},
  year={2023}
}

@inproceedings{xiang2020sapien,
  title={Sapien: A simulated part-based interactive environment},
  author={Xiang, Fanbo and Qin, Yuzhe and Mo, Kaichun and Xia, Yikuan and Zhu, Hao and Liu, Fangchen and Liu, Minghua and Jiang, Hanxiao and Yuan, Yifu and Wang, He and others},
  booktitle={Proceedings of the IEEE/CVF conference on computer vision and pattern recognition},
  pages={11097--11107},
  year={2020}
}

@misc{sketchfab,
  author       = {Sketchfab},
  title        = {Sketchfab - The leading platform for 3D \& AR on the web},
  howpublished = {\url{https://sketchfab.com/}},
  note         = {Accessed: 2025-05-07}
}

@misc{erqa2024,
  author       = {{Embodied Reasoning}},
  title        = {ERQA: Embodied Reasoning Question Answering Benchmark},
  year         = {2024},
  howpublished = {\url{https://github.com/embodiedreasoning/ERQA}},
  note         = {Accessed: 2025-05-07}
}

@misc{robotverse,
      title={RoboVerse: Towards a Unified Platform, Dataset and Benchmark for Scalable and Generalizable Robot Learning}, 
      author={Haoran Geng and Feishi Wang and Songlin Wei and Yuyang Li and Bangjun Wang and Boshi An and Charlie Tianyue Cheng and Haozhe Lou and Peihao Li and Yen-Jen Wang and Yutong Liang and Dylan Goetting and Chaoyi Xu and Haozhe Chen and Yuxi Qian and Yiran Geng and Jiageng Mao and Weikang Wan and Mingtong Zhang and Jiangran Lyu and Siheng Zhao and Jiazhao Zhang and Jialiang Zhang and Chengyang Zhao and Haoran Lu and Yufei Ding and Ran Gong and Yuran Wang and Yuxuan Kuang and Ruihai Wu and Baoxiong Jia and Carlo Sferrazza and Hao Dong and Siyuan Huang and Yue Wang and Jitendra Malik and Pieter Abbeel},
      year={2025},
      eprint={2504.18904},
      archivePrefix={arXiv},
      primaryClass={cs.RO},
      url={https://arxiv.org/abs/2504.18904}, 
}

@misc{arnold,
      title={ARNOLD: A Benchmark for Language-Grounded Task Learning With Continuous States in Realistic 3D Scenes}, 
      author={Ran Gong and Jiangyong Huang and Yizhou Zhao and Haoran Geng and Xiaofeng Gao and Qingyang Wu and Wensi Ai and Ziheng Zhou and Demetri Terzopoulos and Song-Chun Zhu and Baoxiong Jia and Siyuan Huang},
      year={2023},
      eprint={2304.04321},
      archivePrefix={arXiv},
      primaryClass={cs.AI},
      url={https://arxiv.org/abs/2304.04321}, 
}

@misc{instruction_consistency,
  title = {Task {{Reconstruction}} and {{Extrapolation}} for \${$\pi\_$}0\$ Using {{Text Latent}}},
  author = {Li, Quanyi},
  year = {2025},
  month = aug,
  number = {arXiv:2505.03500},
  eprint = {2505.03500},
  primaryclass = {cs},
  publisher = {arXiv},
  doi = {10.48550/arXiv.2505.03500},
  urldate = {2025-09-25},
  archiveprefix = {arXiv},
  langid = {english}
}

\appendix
\newpage
\section{The usage of LLM}
We acknowledge the use of Large Language Models (LLMs) in the preparation of this work in the following capacities:

\textbf{Writing Assistance and Polishing:} LLMs were employed to aid in refining the clarity and coherence of our manuscript. This includes improving sentence structure, enhancing readability, and ensuring consistent academic writing style throughout the paper. All technical content, experimental results, and scientific contributions remain entirely our own work.

\section{Limitations and Future Work}
\label{sec:limitation}
Despite COIN's comprehensive design, several limitations must be acknowledged: (1) our focus on a single robotic platform in static environments fails to capture the full complexity of dynamic real-world scenarios; (2) the absence of dual-arm manipulation tasks that could reveal additional coordination challenges in interactive reasoning.
\begin{figure}
\centering
\includegraphics[width=0.8\linewidth]{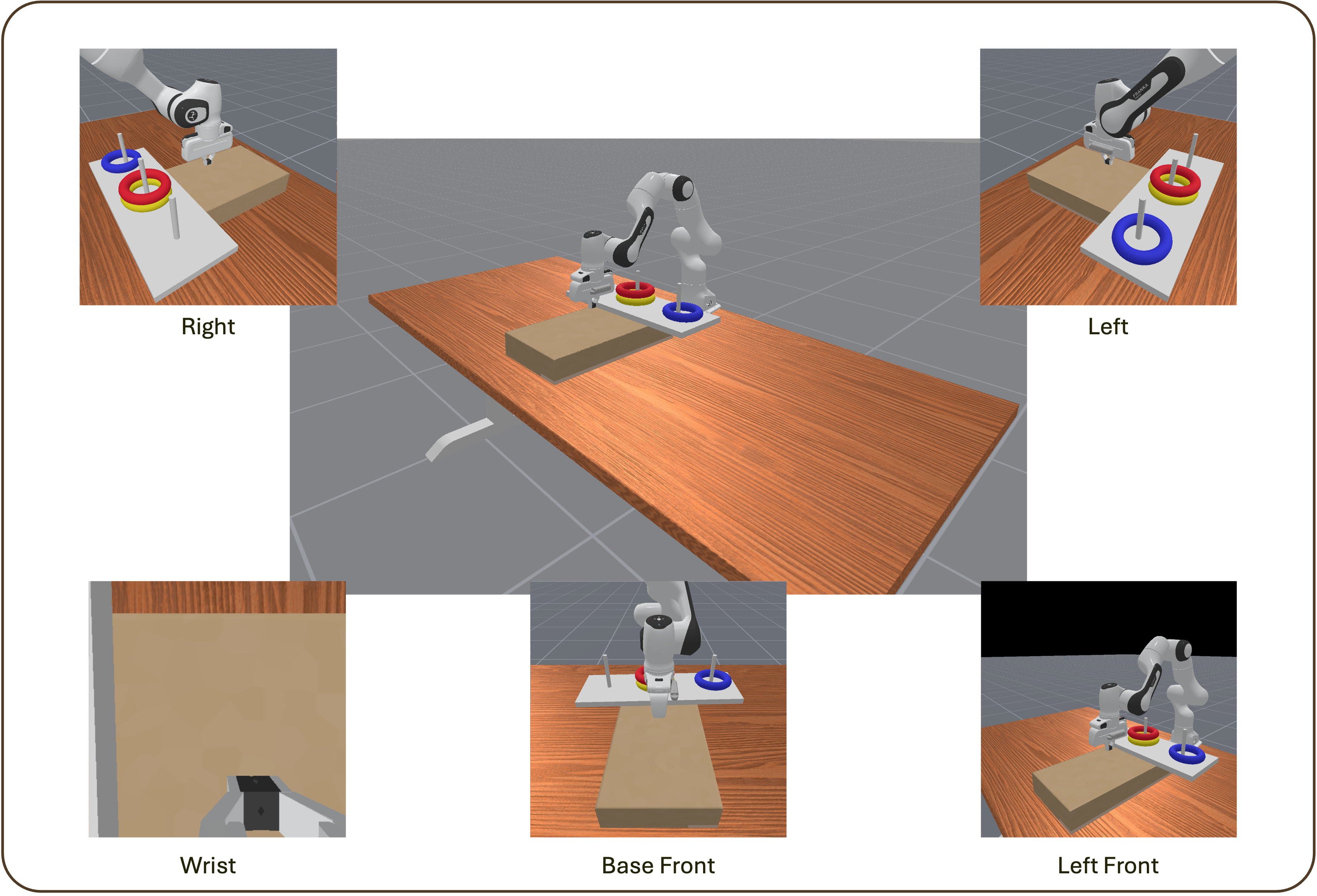}
\caption{Environment Setup}
\label{fig:env_setup}
\end{figure}

For future work, we plan to pursue the promising research directions identified in our conclusions. Specifically, we aim to investigate: (1) trajectory smoothness mechanisms inspired by CogACT's temporal ensemble approach to improve VLA execution stability; (2) enhanced multimodal perception architectures that better integrate visual understanding with instruction following; (3) novel VLM-VLA integration paradigms, comparing latent vector bridges against natural language interfaces; and (4) adaptive CodeAsPolicy frameworks incorporating closed-loop feedback for dynamic re-planning. Additionally, we anticipate significant advancement in human-inspired learning approaches that enable iterative "think-execute-think" cycles, allowing models to formulate hypotheses, design informative tests, and recursively update their world models based on interactive outcomes.

\section{Environment Setup}
\label{sec:env_setup}

We use a 7-DoF Franka Emika Panda robotic arm equipped with a parallel gripper as our standard platform. As described in Section~\ref{sec:formulation}, we use different action spaces for VLA-based controllers and VLM-based planners. Specifically, for VLA models using Panda inverse kinematics, we define the action space as $\Delta p \in [-0.3, 0.3]$ for positional deltas and $\Delta R \in [-0.5, 0.5]$ for orientation deltas. The robot is observed from five camera perspectives providing comprehensive spatial and task context for both VLA and VLM-based planning models. All environments are built on the ManiSkill3 platform using physics-based simulation.

\section{Performance Comparison: Models vs. Human Baseline on COIN-50}

\label{sec:human_baseline}
To contextualize the difficulty level of COIN tasks and establish a performance baseline, we evaluated both current AI systems and human performance on COIN-50 interactive reasoning tasks. For human evaluation, we recruited 3 participants with B.S. degrees who had no prior exposure to our tasks, with each participant attempting a representative subset of 10 tasks twice via teleoperation.

\begin{table}[t]
	\caption{Performance comparison on COIN-50 interactive reasoning tasks. Human participants achieve 40\% success rate via teleoperation (100\% in real-world settings), while current AI approaches rarely exceed 3\% success rates, revealing a substantial capability gap that highlights the significant challenges in achieving interactive reasoning.}
	\label{tab:interactive-success}
	\centering
	\small
	\begin{tabular}{lcccc}
	\toprule
	\textbf{Model/Human} & \textbf{Object-centric} & \textbf{Robot-centric} & \textbf{Compositional} & \textbf{Overall} \\
	\midrule
	\textbf{Human (Sim/  10)} & \multicolumn{3}{c}{N/A} & \textbf{40\%} \\
	\textbf{Human (Real/ 10)} & \multicolumn{3}{c}{N/A} & \textbf{100\%} \\
	\midrule
	\textbf{H-VLA} &  &  &  & \\
	\midrule
	Pi0 + Gemini 2.0 & 1.88\% & 2.14\% & 1.97\% & 1.99\% \\
	Pi0 + GPT-4o & 1.96\% & 2.50\% & 2.05\% & 2.17\% \\
	\midrule
	Gr00t N1 + Gemini 2.0 & 1.74\% & 2.50\% & 1.82\% & 2.02\% \\
	Gr00t N1 + GPT-4o & 1.52\% & 1.79\% & 1.59\% & 1.63\% \\
	\midrule
	CogACT + Gemini 2.0 & 2.14\% & 1.37\% & 2.24\% & 1.92\% \\
	CogACT + GPT-4o & 1.74\% & 1.07\% & 1.82\% & 1.54\% \\
	\midrule
	\textbf{CodeAsPolicy} &  &  &  & \\
	\midrule
	Voxposer(TD) & 0.43\% & 0.00\% & 0.45\% & 0.29\% \\
	Voxposer(Normal) & 2.17\% & 3.57\% & 2.27\% & 2.67\% \\
	Rekep & 3.04\% & 3.57\% & 3.18\% & 3.26\% \\
	\bottomrule
	\end{tabular}
	\end{table}

The results reveal a dramatic performance gap between human capabilities and current AI systems. While humans achieve 40\% success rates in simulation (100\% in real-world settings, confirming task feasibility), the best-performing AI model (Rekep) achieves only 3.26\% overall success rate. This 12-fold performance difference demonstrates that current approaches are fundamentally limited in their interactive reasoning capabilities, with substantial gaps that must be addressed before these systems can effectively operate in partially observable environments requiring adaptive manipulation strategies.

\section{VLM comparison}
\subsection{VLM Planning Comparison}
\label{sec:vlm_compare}
\begin{figure}[!ht]
\centering
\includegraphics[width=0.9\linewidth]{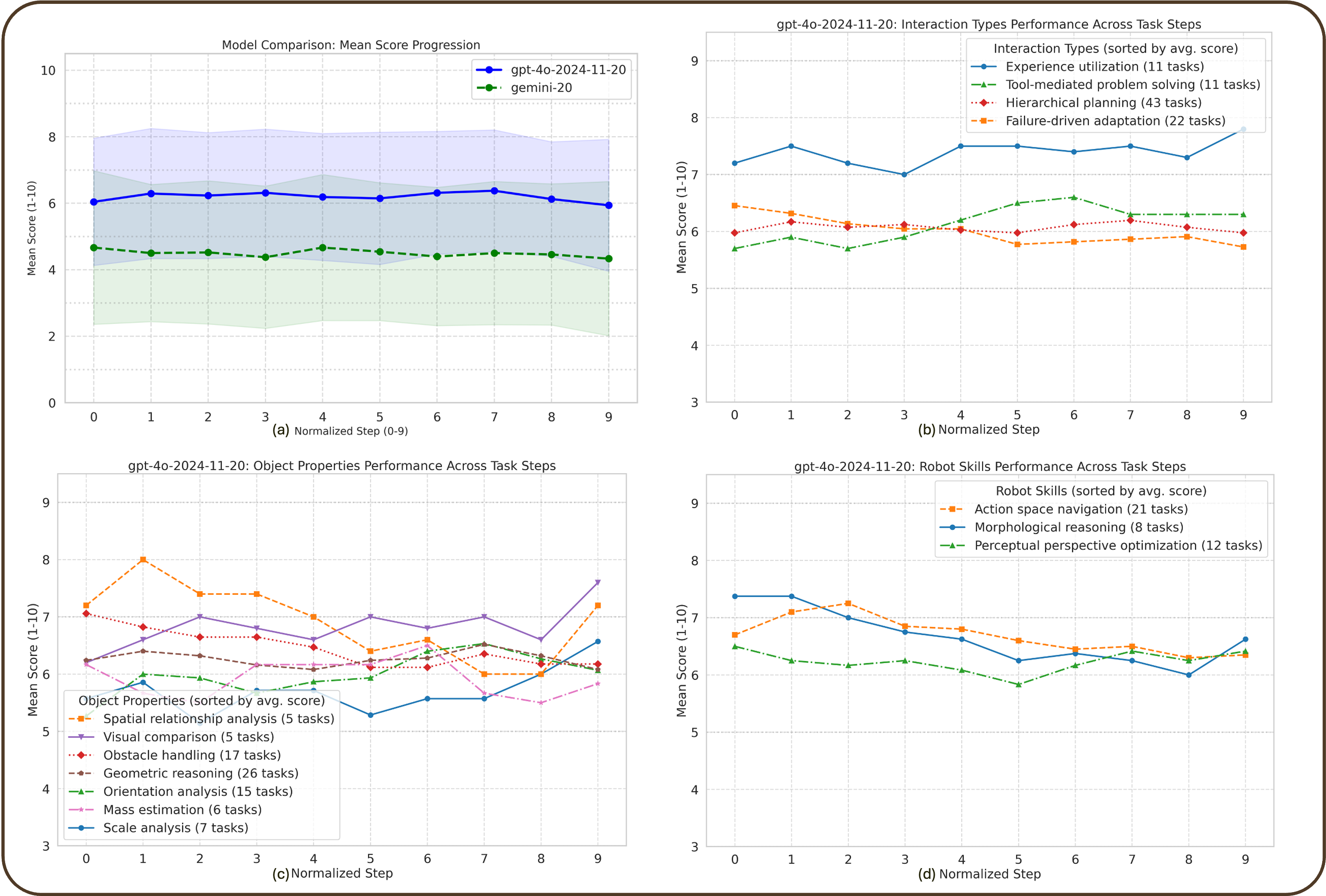}
\caption{Comparison of VLM reasoning abilities on COIN-50 tasks evaluated along expert demonstration videos. GPT-4o consistently outperforms Gemini 2.0 across different reasoning categories and time steps.}
\label{fig:vlm_compare}
\vspace{-3mm}
\end{figure}
We evaluated VLM planners on COIN-50 tasks across execution time steps. GPT-4o consistently outperforms Gemini 2.0 by approximately 1.5 points on our reasoning scale, maintaining this advantage throughout task execution. 
While they could not improve the performance along the time goes on, indicating there might be some problem on models' historical information utilization. 

\subsection{VQA Evaluation Details}
\label{fig:vqa_scores}
While the end-to-end task execution success rates are low, we evaluate the reasoning capabilities of different VLMs through our embodied VQA protocol. The results reveal that GPT-4o consistently outperforms Gemini-2.0-Flash, and all models show increased accuracy in the middle phases of task execution before slightly declining in the final phase. This pattern suggests models gradually accumulate task-relevant information through observation.

\section{Model Training Configurations}
\label{subsec:model-configs}
We use the following training configurations: CogACT-Base trained on 4 $\times$ A800 GPUs with device batch size 32 for 30K steps; Gr00t N1 2B trained on 4 $\times$ A800 GPUs with device batch size 16 for 120K steps; Pi0-Fast trained on 3 $\times$ A800 GPUs with device batch size 2 for 470K steps.

\section{COIN-Teleoperation Algorithm and comparison}

\label{sec:teleoperation}
Our AR teleoperation system demonstrates robust performance across multiple validation metrics:

\textbf{Data Quality Validation:} 90\% of collected trajectories can be successfully replayed in ManiSkill3, indicating high fidelity data capture. Models trained on our data achieve consistent task performance, validating data quality.

\textbf{Cross-Device Compatibility:} The system works reliably across Android and iOS devices released after 2016, with stable 20Hz control frequency maintained even on older hardware (iPhone 7 Plus tested).

\textbf{Comparison with Traditional Methods:} Our approach offers significant advantages in accessibility (no specialized hardware required), scalability (easy deployment across environments), and cost-effectiveness (hardware cost under \$20 vs. thousands for traditional systems).

\begin{algorithm}
\caption{COIN-teleoperation Pipeline}
\begin{algorithmic}[1]
\Require Mobile device with sensors and AR framework
\Ensure Robot control commands

\Function{MobilePhoneProcessing}{}
  \State $X^{IMU}_t \leftarrow$ IMU sensor readings at time $t$
  \State $X^{gyro}_t \leftarrow$ Gyroscope readings at time $t$
  \State $I_t \leftarrow$ Camera image at time $t$
  \State $(p_t, R_t) \leftarrow$ AR framework (ARKit/ARCore) processing of $(X^{IMU}_t, X^{gyro}_t, I_t)$
  \State Establish Web socket connection with PC
  \State Transmit $(p_t, R_t)$ to PC
\EndFunction

\Function{PCProcessingAndControl}{}
  \State Receive $(p_t, R_t)$ from mobile device
  \State $\Delta p_t = p_t - p_{t-1}$, $\Delta R_t = R_t \cdot R_{t-1}^{-1}$
  \State $\Delta \hat{p}_t = \mathrm{MedianFilter}(\{\Delta p_{t-9}, \ldots, \Delta p_t\})$
  \State $\Delta \hat{R}_t = \mathrm{MedianFilter}(\{\Delta R_{t-9}, \ldots, \Delta R_t\})$
  \State $\Delta q_t = \mathrm{InverseKinematics}(\Delta \hat{p}_t, \Delta \hat{R}_t)$
  \State Send joint position commands $\Delta q_t$ to robot
\EndFunction
\end{algorithmic}
\end{algorithm}

\section{More Details about Benchmark}

\subsection{Task Diversity and Temporal Analysis}
\label{sec:temporal_analysis}
COIN demonstrates unprecedented temporal complexity compared to existing benchmarks:

\begin{table}[h]
\centering
\begin{tabular}{lc}
\toprule
\textbf{Benchmark} & \textbf{Average Length} \\
\midrule
ManiSkill & 52.3 \\
CALVIN & 30 \\
Libero & 77.3 \\
ARNOLD & 125.8 \\
VLABench & 157.2 \\
RLBench & 180.2 \\
RoboCASA Composition & 371.9 \\
\textbf{COIN-50} & \textbf{988.9} \\
\bottomrule
\end{tabular}
\caption{Trajectory length comparison across benchmarks. COIN features substantially longer temporal horizons, requiring extended reasoning and planning capabilities.}
\end{table}

\textbf{Multi-Solution Task Diversity:} Over 50\% of COIN tasks exhibit substantial procedural diversity with multiple valid solution paths. For example, in Tabletop-Find-Dice, agents can either systematically examine all faces or directly place dice on markers. This diversity prevents simple memorization and requires genuine reasoning capabilities.

\textbf{Temporal Dependencies:} At least 40 tasks exhibit strong temporal dependencies where earlier information and actions are essential for later stages, ensuring models must reason over extended horizons beyond local cues.

\subsection{Task Classification Details}
\label{subsec:task-class}

COIN evaluates interactive reasoning across three principal domains—object-centric, robot-centric, and compositional—each capturing distinct yet interdependent aspects of embodied intelligence required for manipulation under partial observability.

\subsection{Object-Centric Reasoning}
Object-centric reasoning encompasses an agent's capacity to infer and utilize knowledge about environmental entities through strategic interaction:

\begin{itemize}
	\item \textbf{Physical Property Inference (MAS, FRI, SCA, MOV)}: This category examines the agent's ability to:
	\begin{itemize}
		\item \textit{Mass Estimation (MAS)}: Infer object mass distributions and leverage this information to modulate manipulation forces appropriately for optimal handling.
		\item \textit{Friction Coefficient Assessment (FRI)}: Deduce surface characteristics and their implications for stable grasping and precise manipulation without slippage.
		\item \textit{Scale Analysis (SCA)}: Assess dimensional compatibility between objects and end-effectors to determine feasible manipulation strategies in space-constrained environments.
		\item \textit{Moveable Analysis (MOV)}: Determine which objects or object parts can be manipulated and the constraints on their movement, distinguishing between fixed, partially constrained, and freely movable elements.
	\end{itemize}

	\item \textbf{Spatial Reasoning (OBS, GEO, ORI, SRA)}: This domain evaluates the agent's capacity to:
	\begin{itemize}
		\item \textit{Obstacle Handling (OBS)}: Identify which objects constitute impediments in the manipulation space, distinguish between movable and fixed obstacles, and determine whether to relocate obstacles or navigate around them based on task constraints and efficiency.
		\item \textit{Orientation Analysis (ORI)}: Determine optimal object reorientation for task completion based on geometric and functional constraints in three-dimensional space.
		\item \textit{Spatial Relationship Analysis (SRA)}: Identify and reason about relative positions between objects, including containment relationships (objects inside other objects), proximity relationships (nearest/farthest objects), and spatial arrangements crucial for task execution.
	\end{itemize}

	\item \textbf{Mechanism Understanding (LOC, KIN, SEQ)}: This category evaluates the agent's ability to:
	\begin{itemize}
		\item \textit{Locking System Comprehension (LOC)}: Deduce the operational principles of locking mechanisms and develop appropriate manipulation sequences to engage or disengage them.
		\item \textit{Kinematic Constraint Inference (KIN)}: Identify axes of motion in articulated objects to facilitate effective manipulation within their degrees of freedom.
		\item \textit{Sequential Mechanism Navigation (SEQ)}: Comprehend multi-stage mechanisms and their state-dependent behavior patterns to achieve desired functional outcomes.
	\end{itemize}

	\item \textbf{Visual Reasoning (GEO, VCP, SEM, OCC)}: This category evaluates the agent's ability to process and interpret visual information for manipulation:
	\begin{itemize}
		\item \textit{Geometric Reasoning (GEO)}: Infer shape-based affordances and constraints that influence grasp planning and execution for objects with complex geometries and non-standard forms.
		\item \textit{Visual Comparison (VCP)}: Compare visual properties across object states or between observed objects and internal representations to detect changes, identify matching features, or recognize anomalies.
		\item \textit{Semantic Segmentation (SEM)}: Distinguish between different objects or object parts based on visual features, enabling precise targeting of specific components during manipulation tasks.
		\item \textit{Occlusion Handling (OCC)}: Reason about partially visible or temporarily hidden objects by maintaining object permanence and inferring obscured geometries from limited viewpoints.
	\end{itemize}
\end{itemize}

\subsection{Robot-Centric Reasoning}
Robot-centric reasoning evaluates an agent's capacity for self-awareness and adaptation within the manipulation context, addressing the embodied nature of interaction:

\begin{itemize}
	\item \textbf{Embodiment Awareness (MOR, PPO, KCA)}: This domain assesses the agent's ability to:
	\begin{itemize}
		\item \textit{Morphological Reasoning (MOR)}: Account for the robot's physical dimensions when planning trajectories and interactions to avoid self-collisions and ensure manipulability.
		\item \textit{Perceptual Perspective Optimization (PPO)}: Strategically adjust sensor positioning to maximize information gain during task execution, particularly in partially observable environments.
		\item \textit{Kinematic Constraint Awareness (KCA)}: Reason about joint limitations and workspace boundaries during motion planning to ensure executable action sequences.
	\end{itemize}

	\item \textbf{Control Optimization (DYN, ACT, SKL)}: This category evaluates the agent's capacity to:
	\begin{itemize}
		\item \textit{Dynamic Response Tuning (DYN)}: Adapt control parameters based on task requirements and environmental conditions to achieve desired manipulation outcomes.
		\item \textit{Action Space Navigation (ACT)}: Select appropriate actions from the available repertoire based on current state and task objectives within continuous control spaces.
		\item \textit{Skill Adaptation (SKL)}: Identify and modify learned manipulation skills to meet specific task requirements and environmental conditions.
	\end{itemize}
\end{itemize}

\subsection{Compositional Reasoning Capabilities}
Compositional reasoning encompasses higher-order cognitive functions that integrate multiple reasoning modalities to address complex, interactive challenges requiring adaptive strategies:

\begin{itemize}
	\item \textbf{Tool-Mediated Problem Solving (TOO)}: Identify, create, or adapt tools to overcome manipulation constraints and extend interaction capabilities beyond the robot's native end-effector, enabling solutions to otherwise infeasible tasks.
	\item \textbf{Failure-Driven Adaptation (FDA)}: Actively interact with the environment to gather information about failure modes, then systematically refine strategies based on observed outcomes to develop more robust manipulation approaches through iterative testing.
	\item \textbf{Hierarchical Planning (PLA)}: Decompose complex tasks into coherent sequences of subtasks with appropriate dependencies, adjusting the plan hierarchy in response to changing environmental conditions or task requirements during execution.
	\item \textbf{Experience Utilization (EXP)}: Incorporate historical interaction data into current decision-making processes, applying lessons from previous manipulation attempts to enhance performance in novel but related scenarios through transfer learning.
\end{itemize}
\section{Tasks	Table}
\subsection{Primitive Task}

\renewcommand{\arraystretch}{1.3}
\begin{longtable}{|>{\centering\arraybackslash}p{4cm}|>{\centering\arraybackslash}p{3cm}|>{\centering\arraybackslash}p{5cm}|}
\hline
\textbf{Task ID} & \textbf{Task Image} & \textbf{Description}  \\
\hline
\endfirsthead

\hline
\textbf{Task ID} & \textbf{Task Image} & \textbf{Description}  \\
\hline
\endhead

Tabletop-Close-Cabinet-v1 & \includegraphics[width=2.5cm]{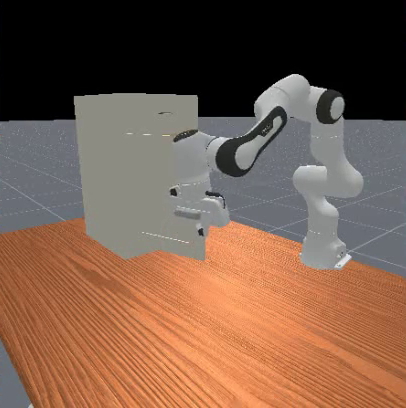} & Close the cabinet door \\[0.3cm] \hline
Tabletop-Close-Door-v1 & \includegraphics[width=2.5cm]{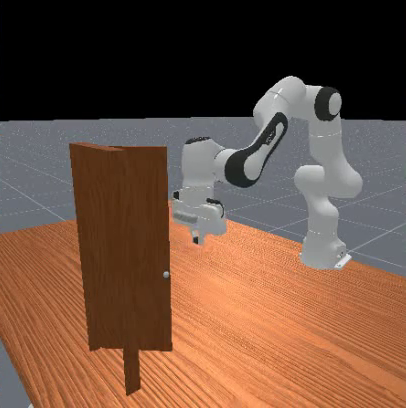} & Close the door \\[0.3cm] \hline
Tabletop-Close-Drawer-v1 & \includegraphics[width=2.5cm]{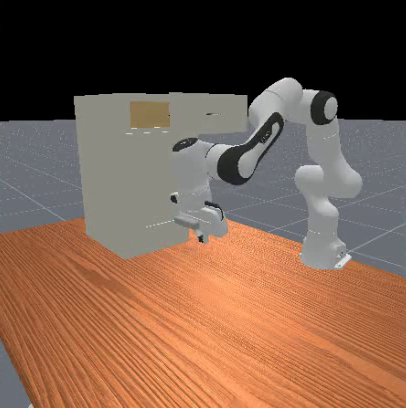} & Close the drawer \\[0.3cm] \hline
Tabletop-Close-Microwave-v1 & \includegraphics[width=2.5cm]{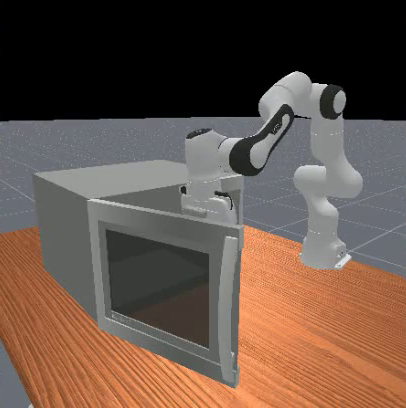} & Close the microwave \\[0.3cm] \hline
Tabletop-Open-Cabinet-v1 & \includegraphics[width=2.5cm]{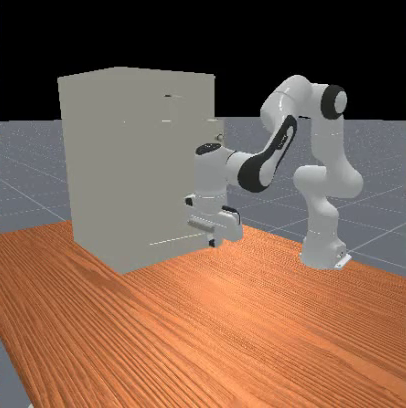} & Open the cabinet door \\[0.3cm] \hline
Tabletop-Open-Door-v1 & \includegraphics[width=2.5cm]{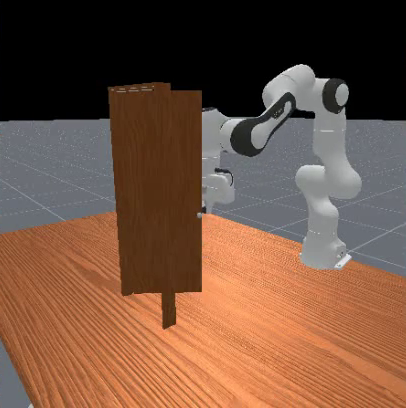} & Open the door \\[0.3cm] \hline
Tabletop-Open-Drawer-v1 & \includegraphics[width=2.5cm]{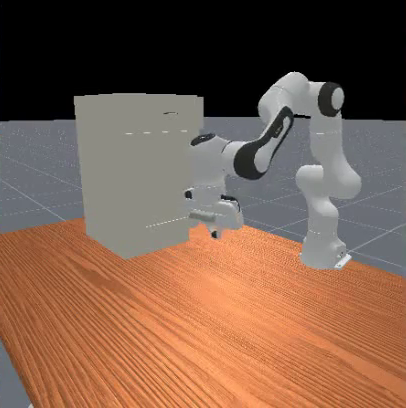} & Open the drawer \\[0.3cm] \hline
Tabletop-Open-Microwave-v1 & \includegraphics[width=2.5cm]{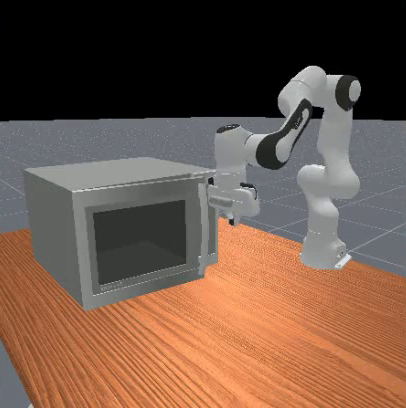} & Open the microwave \\[0.3cm] \hline
Tabletop-Open-Trigger-v1 & \includegraphics[width=2.5cm]{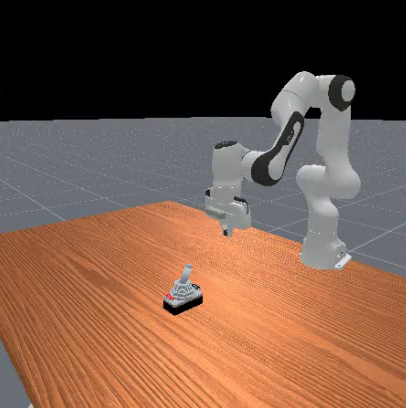} & Turn on the trigger \\[0.3cm] \hline
Tabletop-Pick-Apple-v1 & \includegraphics[width=2.5cm]{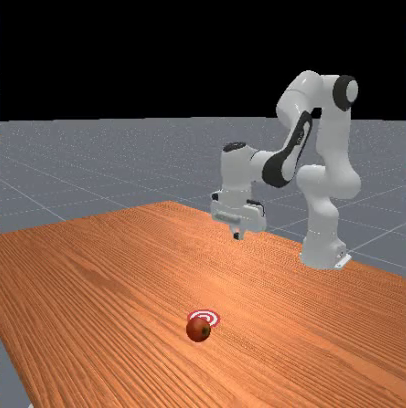} & Pick the apple to the marker \\[0.3cm] \hline
Tabletop-Pick-Book-FromShelf-v1 & \includegraphics[width=2.5cm]{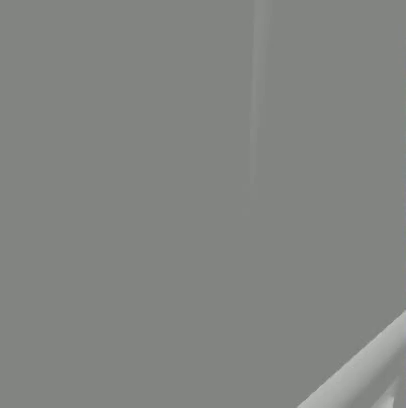} & Find and pick the book from the bookshelf \\[0.3cm] \hline
Tabletop-Pick-Bottle-v1 & \includegraphics[width=2.5cm]{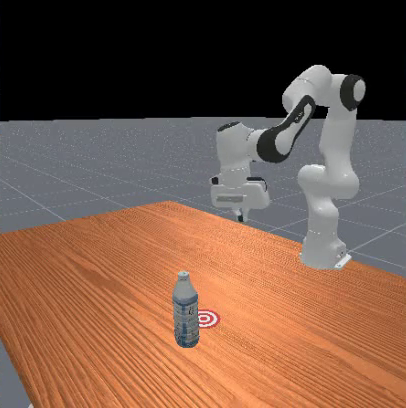} & Pick up the bottle and put it on the marker \\[0.3cm] \hline
Tabletop-Pick-Cube-ToHolder-v1 & \includegraphics[width=2.5cm]{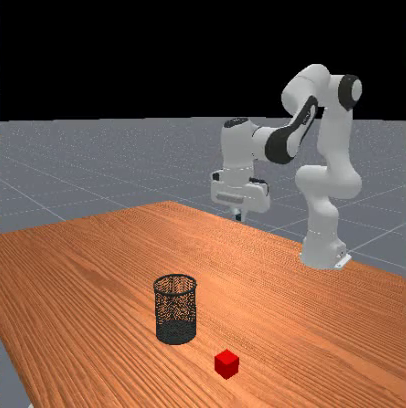} & Pick up the cube, put it in the holder \\[0.3cm] \hline
Tabletop-Pick-Pen-v1 & \includegraphics[width=2.5cm]{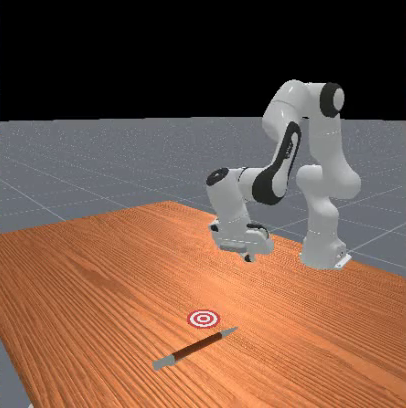} & Pick up the pen and put it to the marker \\[0.3cm] \hline
Tabletop-Pull-Pivot-v1 & \includegraphics[width=2.5cm]{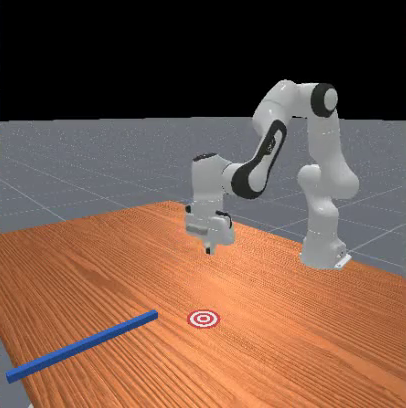} & Pull the pivot to the target area \\[0.3cm] \hline
Tabletop-Put-Ball-IntoContainer-v1 & \includegraphics[width=2.5cm]{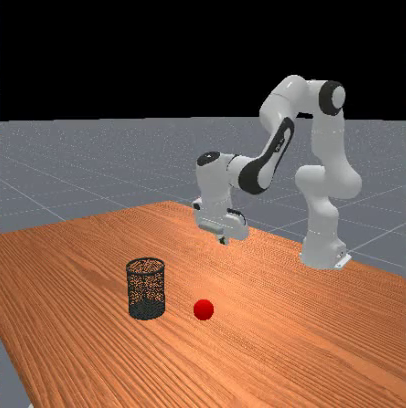} & Put the ball into the container \\[0.3cm] \hline
Tabletop-Put-Fork-OnPlate-v1 & \includegraphics[width=2.5cm]{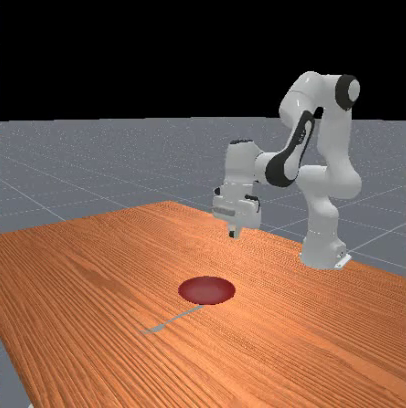} & Put the fork on the plate \\[0.3cm] \hline
Tabletop-Rotate-Cube-v1 & \includegraphics[width=2.5cm]{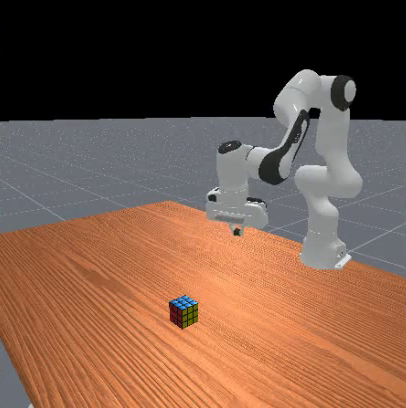} & Rotate the cube till the white face upward \\[0.3cm] \hline
Tabletop-Rotate-Holder-v1 & \includegraphics[width=2.5cm]{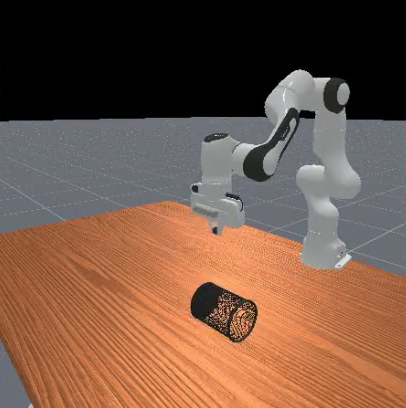} & Rotate the holder till the hole upward \\[0.3cm] \hline
Tabletop-Rotate-USB-v1 & \includegraphics[width=2.5cm]{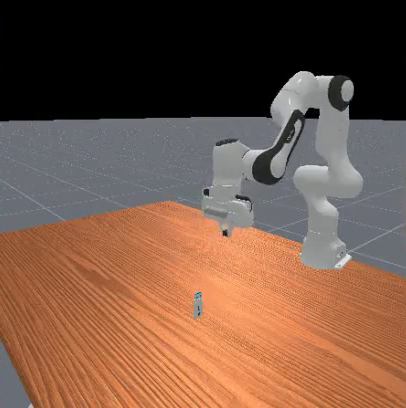} & Rotate the USB body for 90 degrees \\[0.3cm] \hline
Tabletop-Stack-Cubes-v1 & \includegraphics[width=2.5cm]{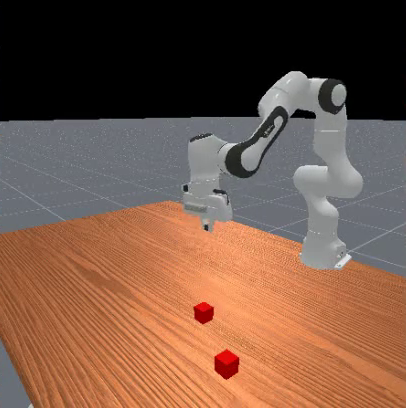} & Stack all the cubes \\[0.3cm]
\bottomrule
\caption{Complete COIN-Primitive task specifications with visual examples (20 tasks)}
\label{tab:coin-primitive-tasks}
\end{longtable}

\subsection{Interactive Task}

\renewcommand{\arraystretch}{1.3}
\begin{longtable}{|>{\centering\arraybackslash}p{2.5cm}|>{\centering\arraybackslash}p{2.5cm}|>{\centering\arraybackslash}p{3.5cm}|>{\centering\arraybackslash}p{1.2cm}|>{\centering\arraybackslash}p{1.2cm}|>{\centering\arraybackslash}p{1.2cm}|}
\hline
\textbf{Task ID} & \textbf{Task Image} & \textbf{Description} & \textbf{Obj.} & \textbf{Rob.} & \textbf{Comp.} \\
\hline
\endfirsthead

\hline
\textbf{Task ID} & \textbf{Task Image} & \textbf{Description} & \textbf{Obj.} & \textbf{Rob.} & \textbf{Comp.} \\
\hline
\endhead
Tabletop-Balance-Pivot-WithBalls-v1 & \includegraphics[width=2.2cm]{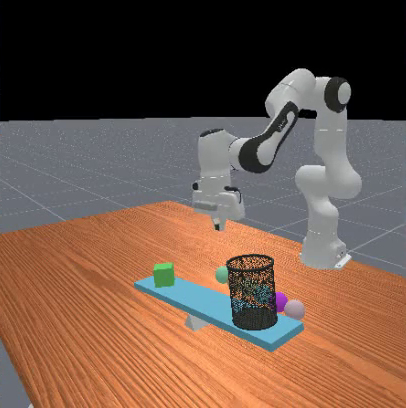} & Put the balls in to the holder to balance the long board on the triangular prism & MAS SCA & no & TOO LPE FDA PLA \\ \hline
Tabletop-Clean-For-Dinner-v1 & \includegraphics[width=2.2cm]{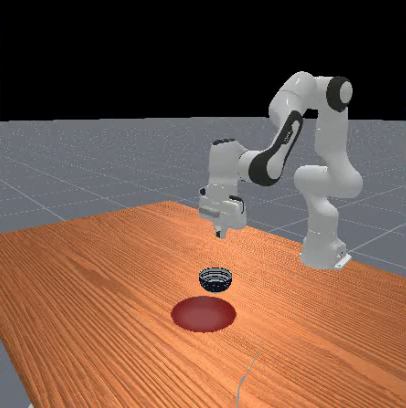} & Arrange the bowl, fork onto the plate, clean for dinner & no & no & no \\ \hline
Tabletop-Close-Cabinet-WithObstacle-v1 & \includegraphics[width=2.2cm]{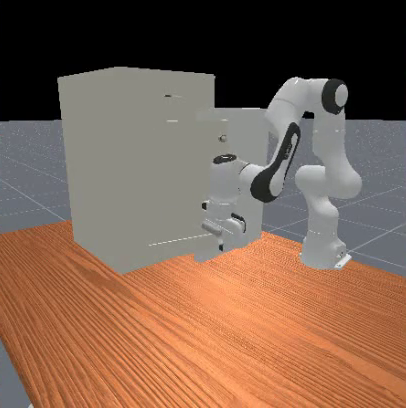} & close the cabinet door & OBS & no & PLA \\ \hline
Tabletop-Close-Door-WithObstacle-v1 & \includegraphics[width=2.2cm]{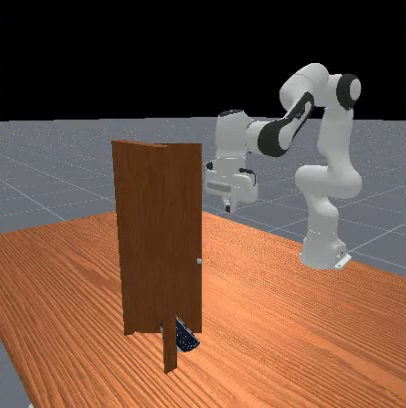} & close the door & no & no & no \\ \hline
Tabletop-Close-Drawer-WithLongObstacle-v1 & \includegraphics[width=2.2cm]{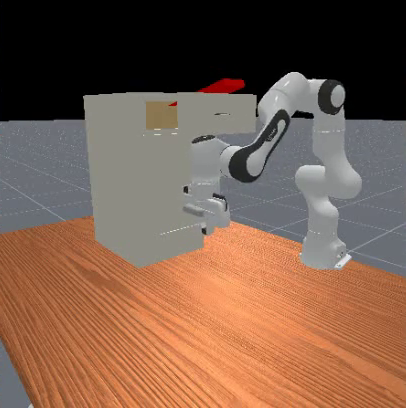} & close the drawer & OBS GEO & no & PLA FDA \\ \hline
Tabletop-Close-Drawer-WithObstacle-v1 & \includegraphics[width=2.2cm]{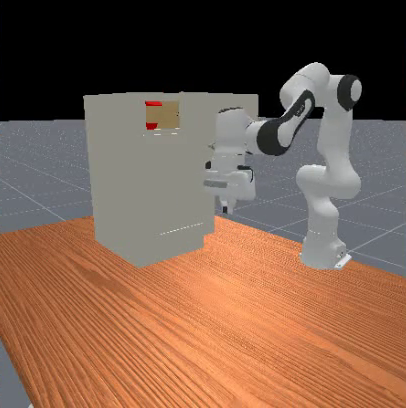} & close the drawer & OBS GEO & ACT PPO & PLA FDA \\ \hline
Tabletop-Find-Book-Black-v1 & \includegraphics[width=2.2cm]{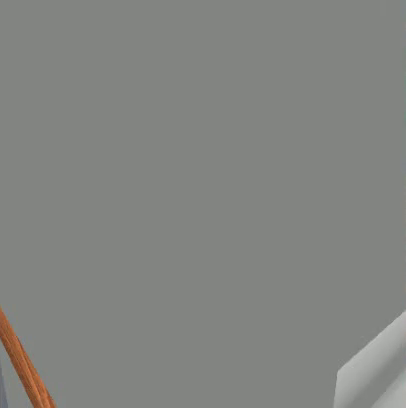} & Find and pick the black book from the bookshelf and put it on the marker & GEO OBS & PPO & EXP \\ \hline
Tabletop-Find-Book-FromShelf-v1 & \includegraphics[width=2.2cm]{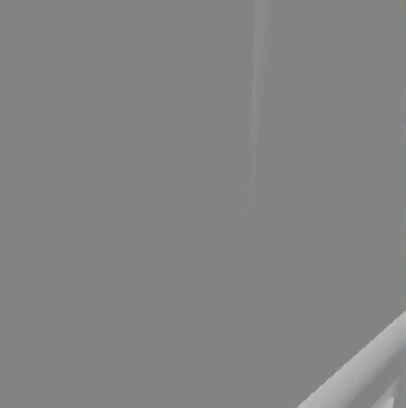} & Find and pick the highest book from the bookshelf and put it on the marker & GEO & PPO & EXP \\ \hline
Tabletop-Find-Cube-RedDown-v1 & \includegraphics[width=2.2cm]{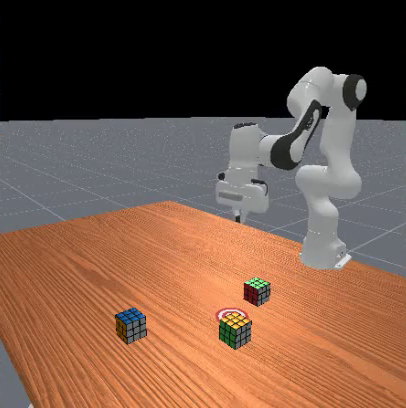} & find the cube which have red face downward, and put it on the marker with red face upward & ORI & PPO ACT & EXP PLA \\ \hline
Tabletop-Find-Cube-WithPivot-v1 & \includegraphics[width=2.2cm]{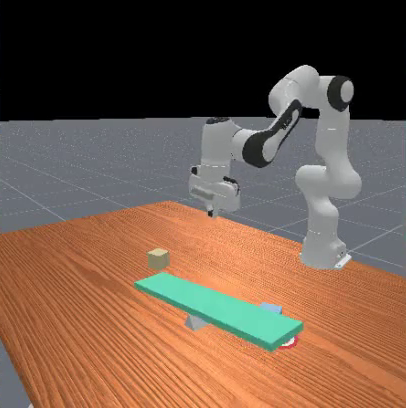} & Move the heavy cube to the goal region & MAS & no & TOO PLA \\ \hline
Tabletop-Find-Dice-v1 & \includegraphics[width=2.2cm]{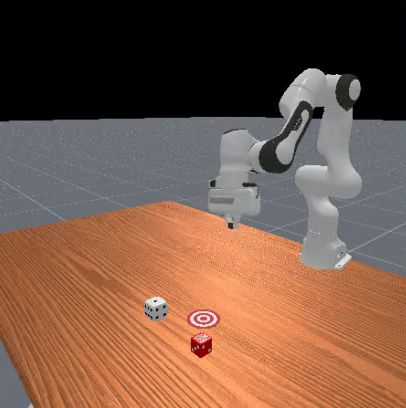} & find the dice which have 2 and 4 point in the corresponding face and put it on the marker & GEO ORI & PPO & EXP PLA \\ \hline
Tabletop-Finish-Hanobi-v1 & \includegraphics[width=2.2cm]{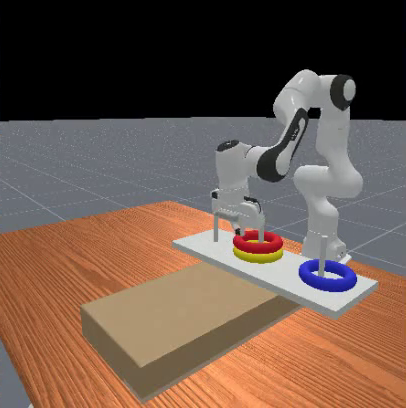} & Place all the hanobi in big to small from bottom to up & GEO SEQ & ACT & PLA EXP \\ \hline
Tabletop-Insert-Conical-v1 & \includegraphics[width=2.2cm]{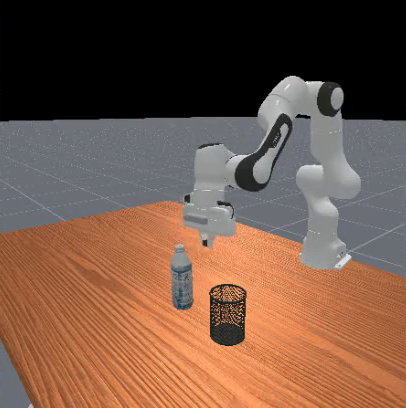} & insert the conical to the container & GEO ORI & no & no \\ \hline
Tabletop-Insert-Objects-WithShape-v1 & \includegraphics[width=2.2cm]{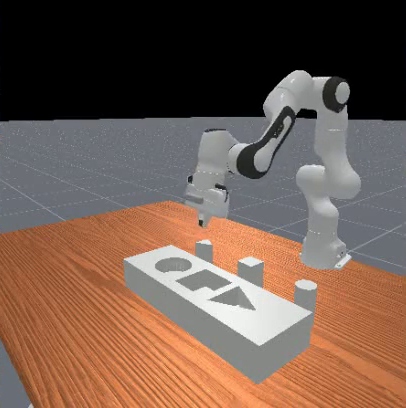} & insert all the stick on the table into corresponding holes & GEO & no & PLA \\ \hline
Tabletop-Insert-WithOrientation-v1 & \includegraphics[width=2.2cm]{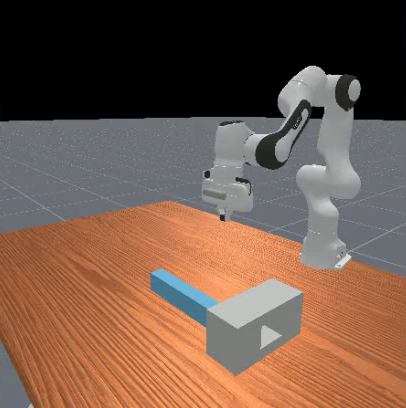} & insert the board on the wall & GEO ORI & no & PLA FDA \\ \hline
Tabletop-Keep-Pivot-Balance-v1 & \includegraphics[width=2.2cm]{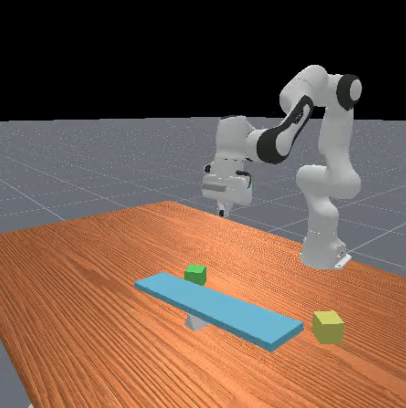} & Balance the long board on the triangular prism and place the cubes to maintain balance & MAS & no & TOO LPE FDA \\ \hline
Tabletop-Lift-Book-v1 & \includegraphics[width=2.2cm]{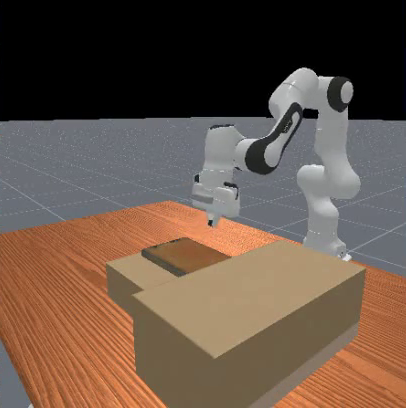} & lift the book up to the higher platform & GEO ORI SCA & MOR & PLA \\ \hline
Tabletop-Merge-Box-v1 & \includegraphics[width=2.2cm]{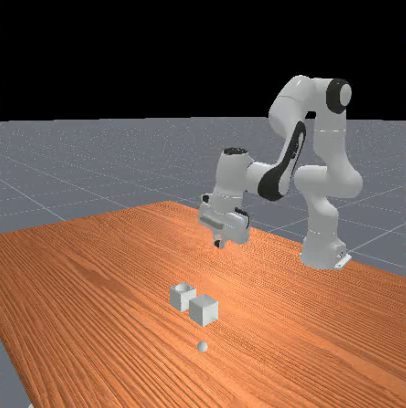} & Merge ball and boxs up & GEO ORI & no & no \\ \hline
Tabletop-Merge-USB-v1 & \includegraphics[width=2.2cm]{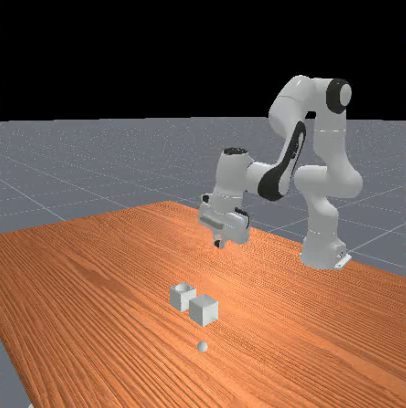} & Pick up the USB body and insert it into the USB hub & GEO & no & PLA \\ \hline
Tabletop-Move-Balls-WithDustpan-v1 & \includegraphics[width=2.2cm]{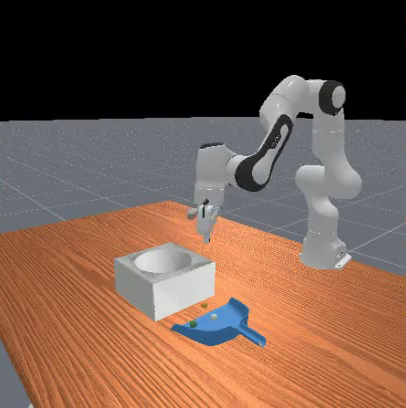} & move all the balls into the holder with dustpan & MAS SCA GEO & no & TOO LPE \\ \hline
Tabletop-Move-Balls-WithPivot-v1 & \includegraphics[width=2.2cm]{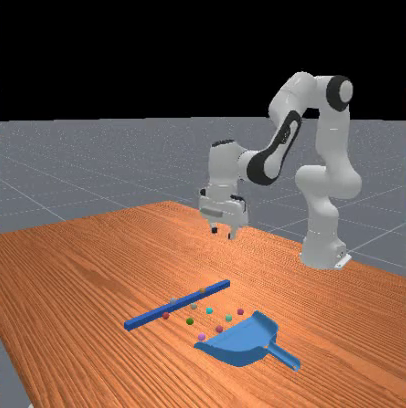} & move all the balls into the dustpan as fast as you can & SCA GEO & no & TOO LPE PLA \\ \hline
Tabletop-Move-Cross-WithStick-v1 & \includegraphics[width=2.2cm]{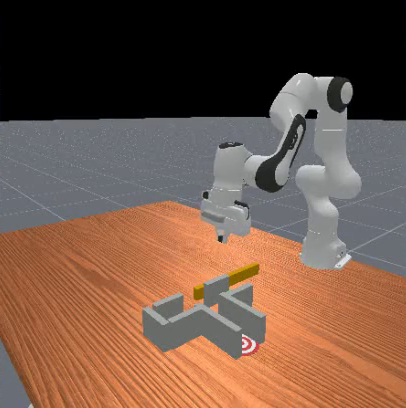} & Use the stick to move the small cube along the cross-shaped path to the target position & no & no & no \\ \hline
Tabletop-Move-Cube-DynamicFriction-v1 & \includegraphics[width=2.2cm]{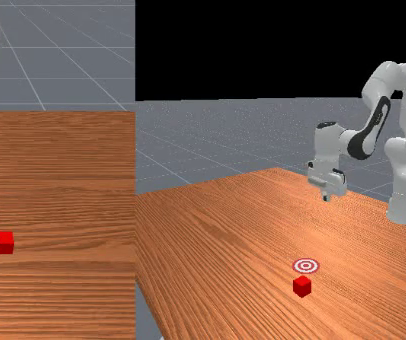} & move the cube to the marker & FRI MAS & no & PLA LPE FDA \\ \hline
Tabletop-Move-Cube-WithHolder-v1 & \includegraphics[width=2.2cm]{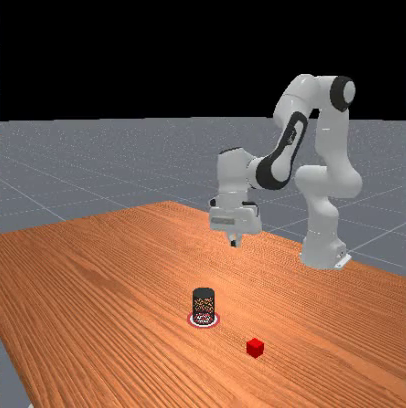} & move the cube to the marker and put the holder on the cube & SCA GEO SEQ & no & PLA \\ \hline
Tabletop-Move-Cube-WithPivot-v1 & \includegraphics[width=2.2cm]{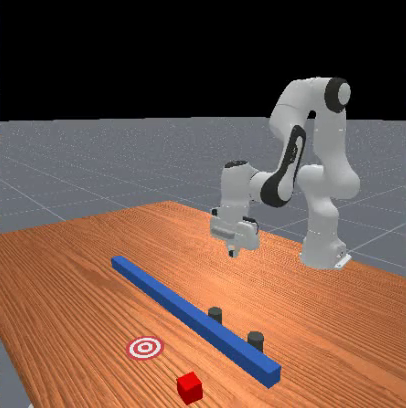} & move the cube with the pivot to the marker & MAS & ACT DYN & PLA TOO LPE FDA \\ \hline
Tabletop-Move-Line-WithStick-v1 & \includegraphics[width=2.2cm]{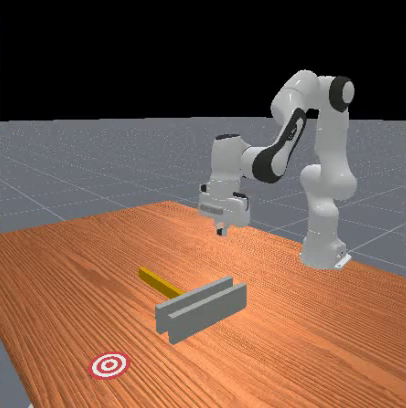} & Use the stick to move the small cube along the straight line path to the target position & GEO ORI & ACT & PLA TOO \\ \hline
Tabletop-Open-Cabinet-WithDoor-v1 & \includegraphics[width=2.2cm]{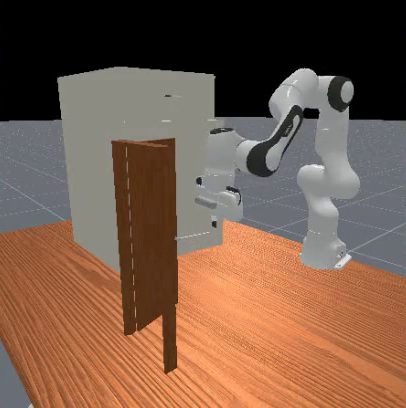} & open the cabinet door & OBS & ACT & PLA \\ \hline
Tabletop-Open-Cabinet-WithObstacle-v1 & \includegraphics[width=2.2cm]{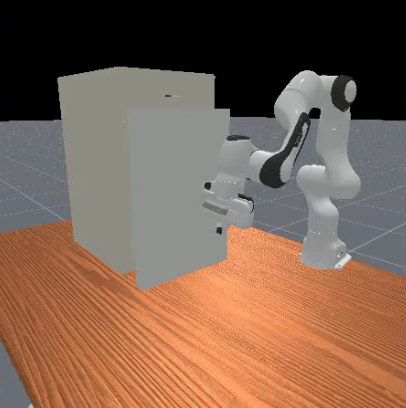} & open the cabinet door & OBS & no & PLA \\ \hline
Tabletop-Open-Cabinet-WithSwitch-v1 & \includegraphics[width=2.2cm]{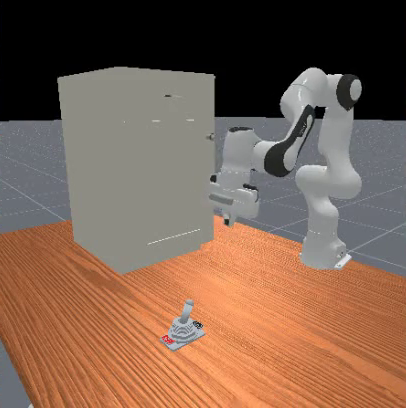} & open the door, notice the switch will control the state of the door & LOC & no & PLA FDA \\ \hline
Tabletop-Open-Door-WithCabinet-v1 & \includegraphics[width=2.2cm]{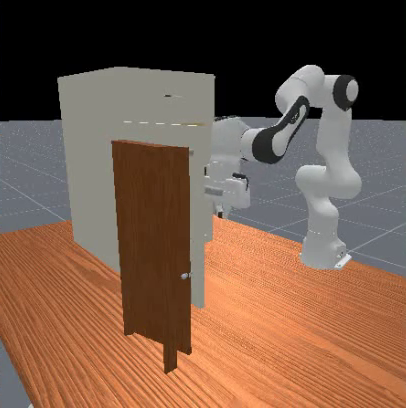} & open the door & OBS & no & PLA \\ \hline
Tabletop-Open-Door-WithObstacle-v1 & \includegraphics[width=2.2cm]{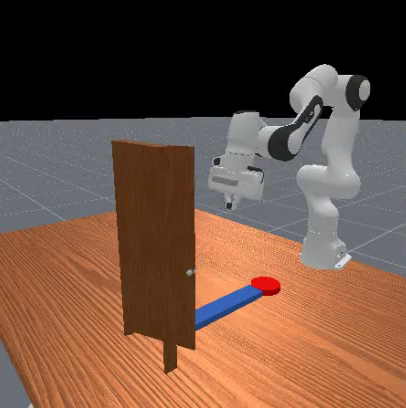} & open the door & OBS & no & PLA FDA \\ \hline
Tabletop-Pick-Cube-Slippery-v1 & \includegraphics[width=2.2cm]{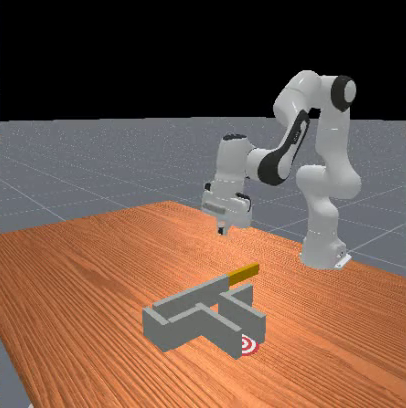} & Pick the slippery cube & FRI & ACT & PLA TOO LPE FDA \\ \hline
Tabletop-Pick-Cube-WithDoor-v1 & \includegraphics[width=2.2cm]{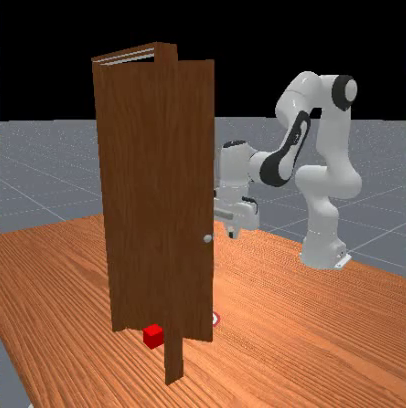} & put the cube to the marker & OBS & KIN ACT & PLA FDA \\ \hline
Tabletop-Pick-Cube-WithStick-v1 & \includegraphics[width=2.2cm]{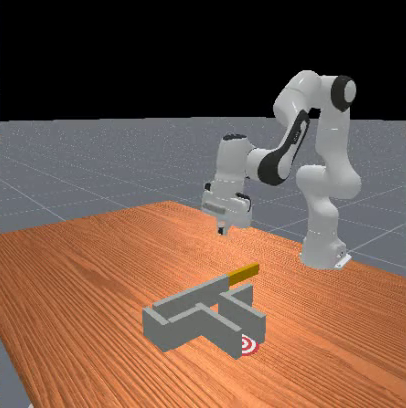} & Use the stick to move the small cube along the T-shaped path to the target position & GEO ORI & ACT & PLA TOO \\ \hline
Tabletop-Pick-Cylinder-WithObstacle-v1 & \includegraphics[width=2.2cm]{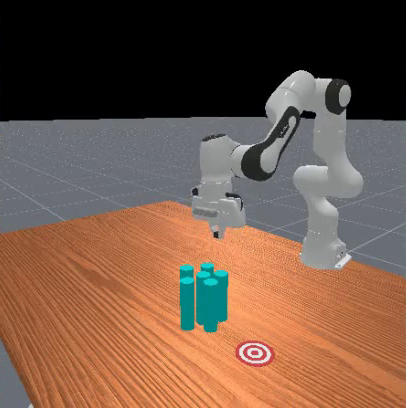} & pick up the center cylinder & LOC & KIN & PLA FDA EXP \\ \hline
Tabletop-Pick-Eraser-FromHolder-v1 & \includegraphics[width=2.2cm]{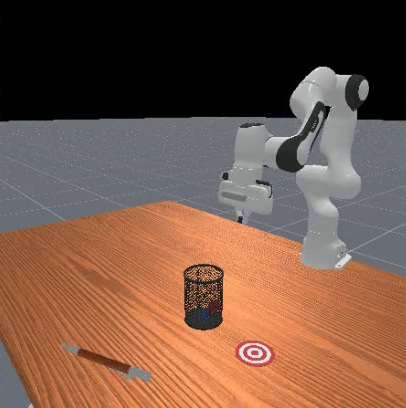} & Pick up the eraser in the holder and place it to the marker & GEO ORI & MOR & PLA EXP \\ \hline
Tabletop-Pick-Object-FromCabinet-v1 & \includegraphics[width=2.2cm]{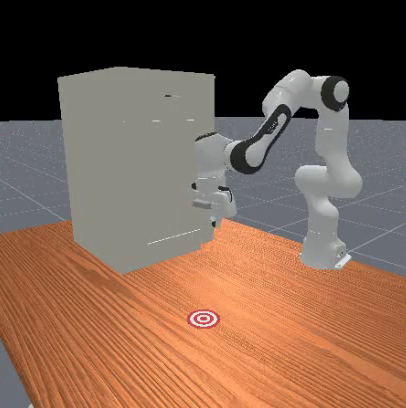} & pick up the object from the cabinet & OBS GEO & PPO MOR ACT & PLA FDA \\ \hline
Tabletop-Put-Balls-IntoContainer-v1 & \includegraphics[width=2.2cm]{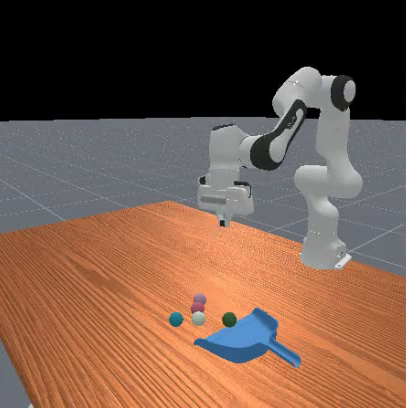} & move all the balls into the dustpan as fast as you can & GEO & ACT & TOO LPE PLA FDA \\ \hline
Tabletop-Put-Cube-IntoCabinetWithObstacle-v1 & \includegraphics[width=2.2cm]{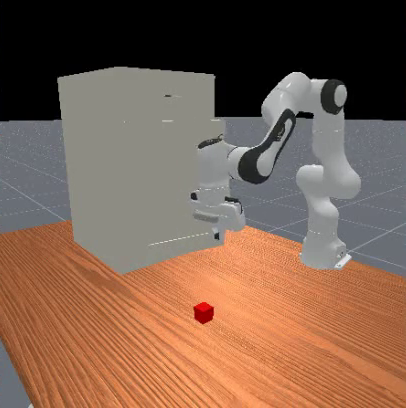} & put the object into the cabinet & OBS GEO & PPO MOR ACT & PLA FDA \\ \hline
Tabletop-Put-Cube-IntoMicrowave-v1 & \includegraphics[width=2.2cm]{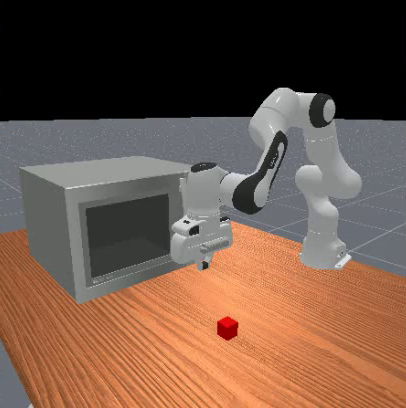} & put the cube into the microwave & OBS GEO & ACT MOR & PLA FDA \\ \hline
Tabletop-Rotate-Cube-Twice-v1 & \includegraphics[width=2.2cm]{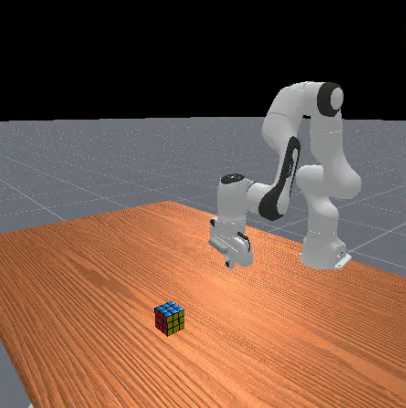} & rotate the cube till the green face upward & ORI & ACT & PLA FDA \\ \hline
Tabletop-Seek-Holder-InCabinet-v1 & \includegraphics[width=2.2cm]{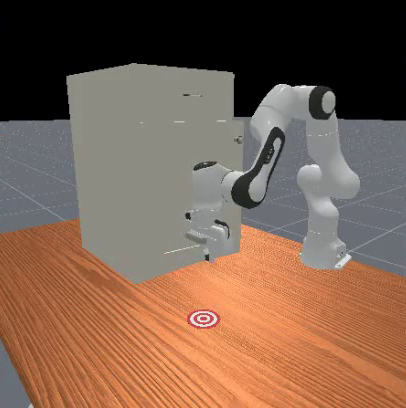} & Find the holder containing an eraser the cabinet, put it to the marker & OBS GEO SEQ & ACT MOR PPO & PLA EXP \\ \hline
Tabletop-Seek-Objects-InCabinet-v1 & \includegraphics[width=2.2cm]{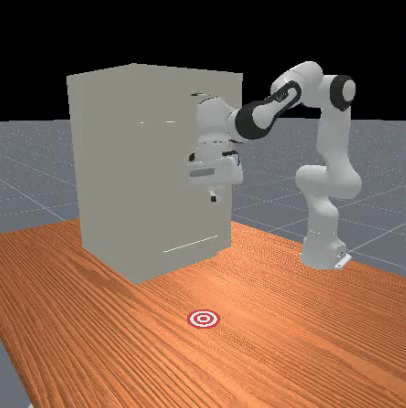} & Find the apple and the plate in the cabinet, put them on the table & OBS & ACT MOR PPO & FDA PLA EXP \\ \hline
Tabletop-Seek-Objects-WithObstacle-v1 & \includegraphics[width=2.2cm]{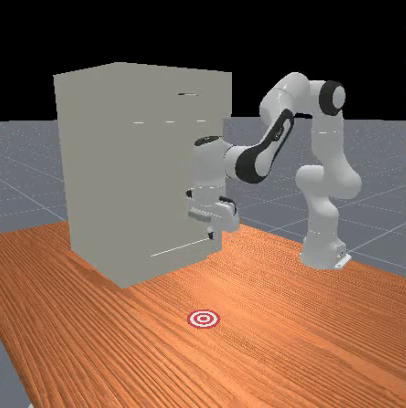} & find the cube in the cabinet and pick it up & OBS & MOR ACT PPO & FDA PLA \\ \hline
Tabletop-Slide-Cube-Into-Container-v1 & \includegraphics[width=2.2cm]{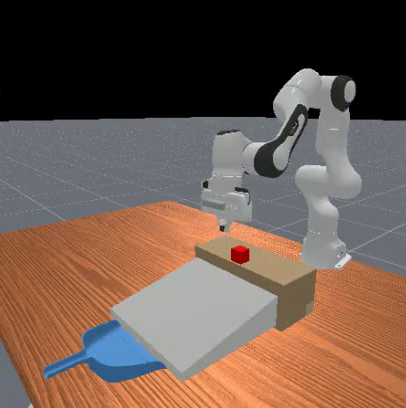} & Slide a cube down a slope into a container & no & no & no \\ \hline
Tabletop-Slide-Cube-WithPath-v1 & \includegraphics[width=2.2cm]{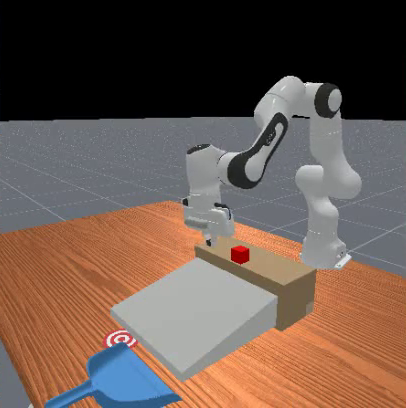} & Slide a cube down a slope to the marker & FRI ORI GEO & ACT & PLA \\ \hline
Tabletop-Stack-Books-OnBox-v1 & \includegraphics[width=2.2cm]{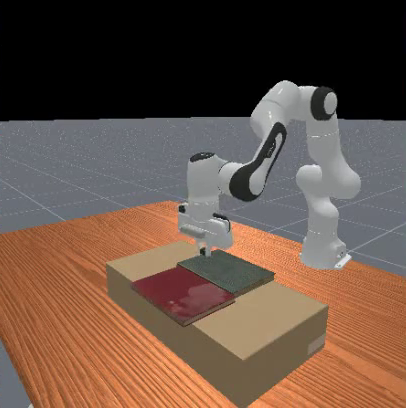} & Stack all things on the table & ORI GEO & PPO & PLA \\ \hline
Tabletop-Stack-Books-v1 & \includegraphics[width=2.2cm]{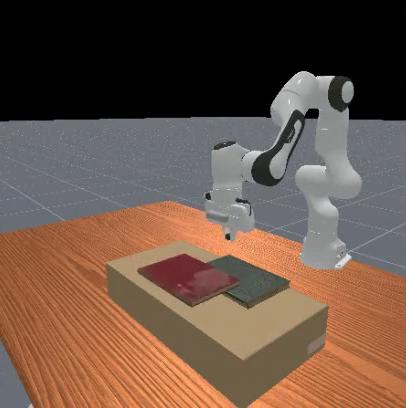} & Stack all things on the table & SCA ORI & PPO & PLA \\ \hline
Tabletop-Stack-Cube-WithColor-v1 & \includegraphics[width=2.2cm]{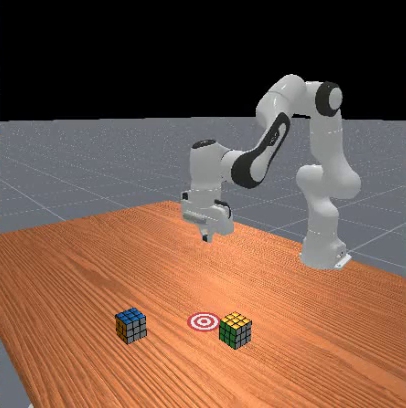} & Stack the cube with same color & ORI & ACT & PLA FDA \\ \hline
Tabletop-Stack-LongObjects-v1 & \includegraphics[width=2.2cm]{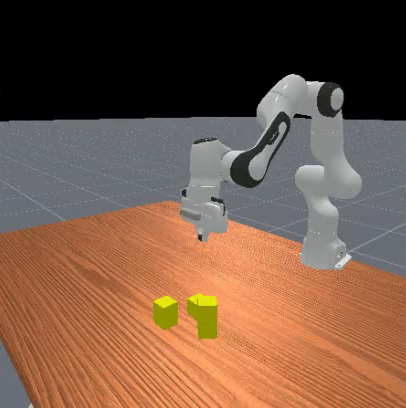} & stack all the objects to make it most high & SCA ORI GEO OBS & ACT & FDA PLA \\ \hline
\bottomrule
\caption{COIN-50 interactive reasoning task specifications with visual examples (50 tasks)}
\label{tab:coin-50-tasks}
\end{longtable}

\end{document}